\documentclass[11pt]{article}

\usepackage[T1]{fontenc}

\usepackage{helvet}
\usepackage[margin=1in]{geometry}
\usepackage{titlesec}
\usepackage{graphicx}
\usepackage{parskip}
\usepackage[dvipsnames]{xcolor}
\usepackage{tcolorbox}
\usepackage[numbers,square]{natbib}  
\usepackage{amsmath, amssymb}
\usepackage{fancyhdr}
\usepackage{microtype}
\usepackage{enumitem}
\usepackage[hidelinks]{hyperref}  
\usepackage[left]{lineno}

\usepackage{url}
\providecommand{\doi}[1]{\url{https://doi.org/#1}}
\usepackage{booktabs}
\usepackage{amsfonts}
\usepackage{nicefrac}
\usepackage{float}
\usepackage{algorithm}
\usepackage{algpseudocode}
\usepackage{inconsolata}
\usepackage[labelfont=bf]{caption}
\usepackage{multirow}
\usepackage{refcount}
\usepackage{footmisc}
\usepackage{wrapfig}
\usepackage{epigraph}

\definecolor{metaBlue}{RGB}{24,119,242}
\definecolor{abstractBoxBG}{RGB}{245,245,250}

\titleformat{\section}{\color{black}\sffamily\Large\bfseries}{\thesection}{0.6em}{}
\titlespacing*{\section}{0pt}{1.5\baselineskip}{1\baselineskip}

\titleformat{\subsection}{\color{black}\sffamily\large\bfseries}{\thesubsection}{0.6em}{}
\titlespacing*{\subsection}{0pt}{1.5\baselineskip}{1\baselineskip}

\usepackage{xspace}
\usepackage{bbm}
\newcommand{\scititle}{$\Psi$-0.6: Adaptive Streams of Thought in a Unified Vision-Language World Model}

\newcommand{\yifan}[1]{#1}  

\usepackage{amsmath,amsfonts,bm}

\def\eqref#1{equation~\ref{#1}}

\def\1{\bm{1}}

\DeclareMathAlphabet{\mathsfit}{\encodingdefault}{\sfdefault}{m}{sl}
\SetMathAlphabet{\mathsfit}{bold}{\encodingdefault}{\sfdefault}{bx}{n}

\begin{document}
\thispagestyle{empty}
\enlargethispage{2\baselineskip}

\newcommand{\changes}[1]{{#1}}
\newcommand{\dataset}[1]{\textsc{#1}}
\newcommand{\dname}{FlowBench}

\vspace*{-2em}
\begin{tcolorbox}[
  colback=abstractBoxBG,
  colframe=abstractBoxBG,  
  boxrule=0pt,
  arc=4pt,
  left=12pt, right=12pt, top=10pt, bottom=12pt,
  width=\textwidth,
  enlarge left by=0mm,
  enlarge right by=0mm
]

{\sffamily\LARGE\bfseries \scititle \\[0.5em]}
{\bfseries Khai Loong Aw\textsuperscript{*}, Klemen Kotar\textsuperscript{*}, Baihan Zhang\textsuperscript{*}, Yifan Liu\textsuperscript{*},\\
Lilian Chen, Atlas Kazemian, Rahul Venkatesh, Daniel Yamins}\\
\small Stanford University\\[1em]
\textbf{Abstract.} Humans reason about the physical world by flexibly combining different forms of inference as a problem unfolds.
Achieving this flexibility is a major open challenge in AI: capabilities often remain separated across specialized models, limiting how they can be composed and jointly adapted to a task.
We introduce $\Psi$-0.6, a unified \emph{vision-language world model} that adaptively composes perception, language, geometry, and simulation within a shared reasoning process we call \emph{streams of thought}.
These streams interleave images, text, and explicit representations of geometry and motion, all of which the model can generate and interpret through a shared token interface.
$\Psi$-0.6 surpasses leading open-weight models zero-shot on multi-view spatial reasoning, with further gains from reasoning through geometric intermediates.
It also supports \emph{counterfactual simulation}: generating and interpreting imagined scenes to determine, for example, whether moving an object would obstruct a pathway.
Reinforcement learning from task feedback jointly optimizes the selection, generation, and use of intermediate representations, improving visual question-answering accuracy with gains that transfer to unseen benchmarks.
$\Psi$-0.6 offers a path toward adaptive physical reasoning -- not only predicting the physical world but also choosing the representations through which to reason.



\end{tcolorbox}

\begingroup
\renewcommand{\thefootnote}{\fnsymbol{footnote}}
\footnotetext[1]{Equal contribution.}
\endgroup

\vspace{-0.5em}

\begin{figure}[t]
\centering

\includegraphics[width=\textwidth]{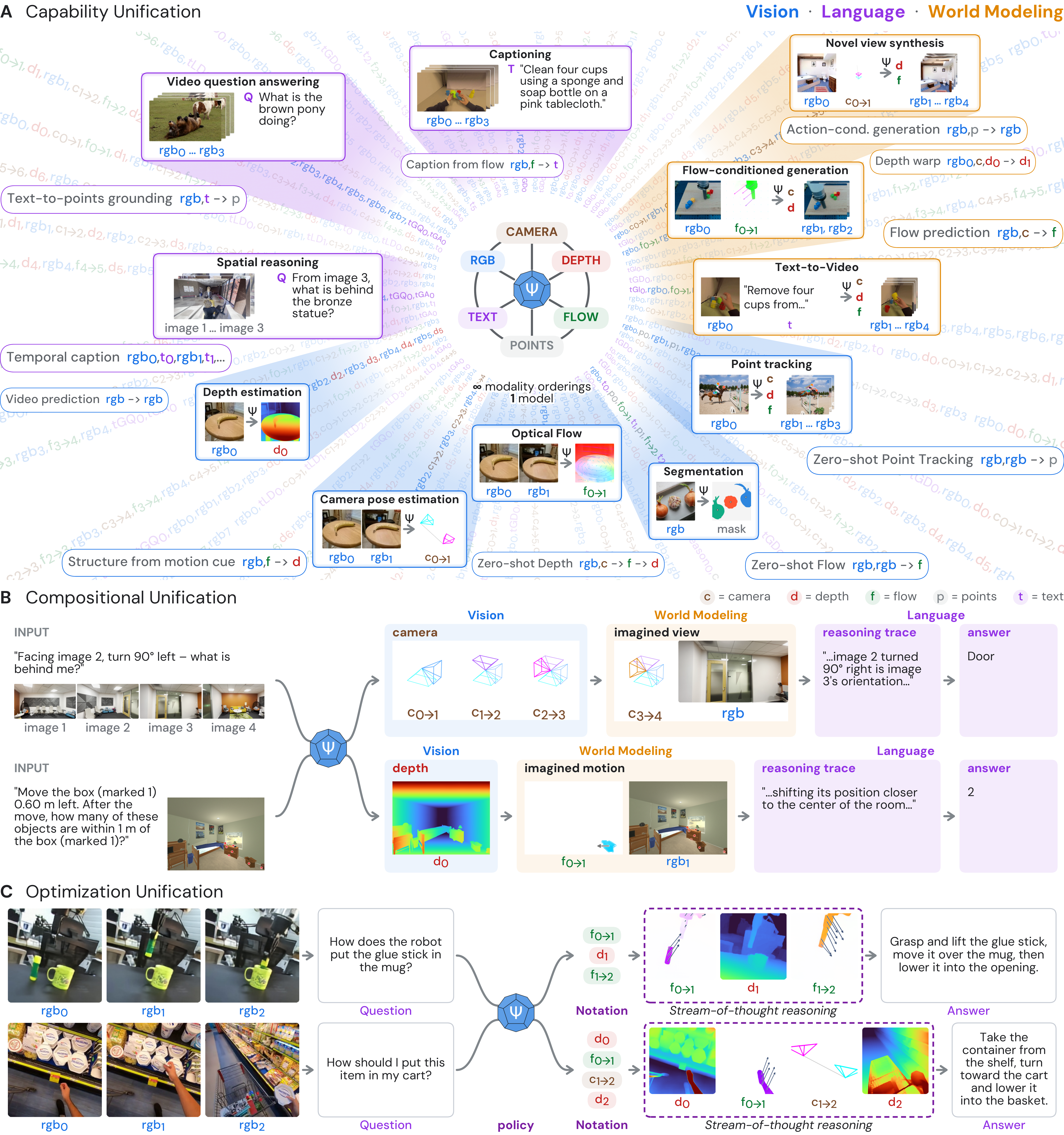}
\caption{
\textbf{$\Psi$-0.6: representation-adaptive reasoning across vision, language, and world modeling.}
\textbf{(A)} \emph{Capability unification.}
Diverse representations share a common token interface. Different choices of supplied and predicted tokens express perception, geometry, language, and imagination within a single transformer.
\textbf{(B)} \emph{Compositional unification.}
Streams of thought combine model capabilities.
\textbf{(C)} \emph{Optimization unification.}
Because intermediate representations are predictions of the same model, choosing which representations to generate and use becomes a learnable action.
}
\label{fig:schematic}
\end{figure}

\section{Introduction}
\label{sec:intro}

\begin{quote}
``Solving a problem simply means representing it so as to make the solution transparent.''
\hfill --- Herbert A. Simon, \textit{The Sciences of the Artificial}
\end{quote}



Humans reason flexibly about the physical world, shifting between different forms of inference as a problem unfolds. When assembling furniture, we read instructions, change viewpoint, and mentally rotate a panel to imagine whether it will fit. These abilities are remarkably powerful, in three related but distinct senses. First, they are \emph{unified}: diverse capabilities are expressed in a single system. Second, they are \emph{compositional}: intermediate inferences and imagined states can be combined into solutions that no single operation provides. Third, they are \emph{adaptive}: which representations are useful depends on the task, and experience can improve how they are selected and used. For example, words express goals and relations; geometric representations expose spatial structure; and imagined scenes reveal possible action consequences \citep{shepard_mental_rotation_1971,battaglia_simulation_2013}. 
Here we ask: {how can an AI system learn which representations are useful for solving problems in the physical world?}

This remains a major challenge for current physical AI systems, as capabilities often remain fragmented across specialized models, limiting how flexibly they can be combined and adapted to a task.
Vision encoders estimate and perceive, vision-language models (VLMs) describe and answer, and world models (WMs) predict and imagine -- each with a separate architecture, training, and interface. This separation makes it difficult to compose these capabilities within a shared reasoning process and jointly optimize how these models' predictions are generated and used through downstream task feedback.
A growing body of work builds unified multimodal models \citep{lu_unified-io_2022, lu_unified-io_2023, bachmann_4m-21_2024,team_chameleon_2025,cui_emu35_2025}, but these capabilities are not integrated into a shared reasoning process. Another line of work allows language or vision-language models to invoke external models, tools, or programs~\citep{schick_toolformer_2023,suris_vipergpt_2023,hu_visual_2024}. However, the invoked components are typically treated as fixed black boxes whose prediction mechanisms are not jointly optimized with the reasoner for downstream tasks.

To achieve flexible reasoning, we turn to a powerful recipe that already exists -- in large language models (LLMs): (i) a unified model provides a broad repertoire of capabilities, (ii) a generic prompting interface makes those capabilities accessible and composable across tasks~\citep{brown_language_2020,kojima_zero_shot_2022}, and (iii) reinforcement learning on task outcomes shapes the intermediate reasoning strategies used to solve a problem \citep{deepseek-ai_deepseek-r1_2025}. Chain-of-thought has been central to the reasoning capabilities of modern LLMs, motivating the question of whether similarly self-directed, adaptive reasoning can extend to the physical world. Doing so requires making visual and geometric predictions available not only as task outputs, but as intermediate operations that a model can generate, compose, and learn to use. 

We introduce $\Psi$-0.6, a unified vision-language world model that adaptively composes perception, language, geometry, and simulation within a shared reasoning process we call \emph{streams of thought}. These are sequences interleaving images and text with explicit representations of geometry and motion (Fig.~\ref{fig:schematic}). $\Psi$-0.6 builds on the \emph{Probabilistic Structure Integration} (PSI; $\Psi$) framework \citep{kotar_world_2025}, extending it to language and flexible reasoning. The model predicts images, text, camera pose, depth, optical flow, and point tracks as tokens with a shared transformer. An {any-to-any} prompting interface allows it to condition on and generate these representations in different combinations, so the output of one operation can become context for the next (\S\ref{sec:prompting}). For example, an image pair can prompt a camera-pose estimate, while an image and camera pose can prompt a new view.

{Unification thus turns representations into reasoning operations}: the same model can generate them, condition on them, compose them, and learn when they are useful. World modeling extends these streams beyond observed scenes through {counterfactual simulation}: asked whether moving an object would obstruct a pathway, the model can translate the specified displacement into an optical-flow prediction, generate the resulting scene, and answer using its own prediction. Crucially, the {reasoner and imaginer are the same model}: predicted scenes are generated and interpreted within the same token interface rather than supplied by a separate visual generator.

Beyond prescribed reasoning pathways, $\Psi$-0.6 learns \emph{representation-adaptive} streams of thought through reinforcement learning on task feedback. Because geometric intermediates, imagined views, and text are all actions of the same policy, downstream reward can shape both which representations are generated and their contents. Unlike VLMs that invoke fixed external tools, $\Psi$-0.6 generates these intermediates within the reasoner itself, enabling their generation and use to be jointly optimized end to end (\S\ref{sec:rl}).

Our experiments demonstrate three benefits of unification: strong zero-shot capabilities, compositional physical reasoning, and the ability to learn adaptive reasoning strategies from task feedback. On MindCube~\citep{wang_mindcube_2026}, a challenging benchmark of multi-view spatial reasoning, $\Psi$-0.6 outperforms leading open-weight models in zero-shot evaluation, with further gains from geometric context and composed streams of thought (\S\ref{sec:mindcube}). On FlowBench~\citep{liu_counterfactual_2026}, when given a specified counterfactual motion, its world-modeling capabilities surpass those of dedicated expert models that receive privileged inputs. In simulation, its performance on counterfactual questions comes close to that of other leading open-source VLMs, and it transfers to real robot data after training only on simulation examples (\S\ref{sec:counterfactual}). Reinforcement learning yields mean relative accuracy gains over a video-only question-answering baseline on held-out data and across four unseen benchmarks, demonstrating transfer of the learned policy (\S\ref{sec:rl}). Together, these results demonstrate our core thesis that unifying perception, geometry, language, and imagination allows task feedback to shape not only what a model predicts but also how it reasons.

\section{A Unified Model with Adaptive Streams of Thought}
\label{sec:method}


\begin{figure}[t]
\centering
\includegraphics[width=\textwidth]{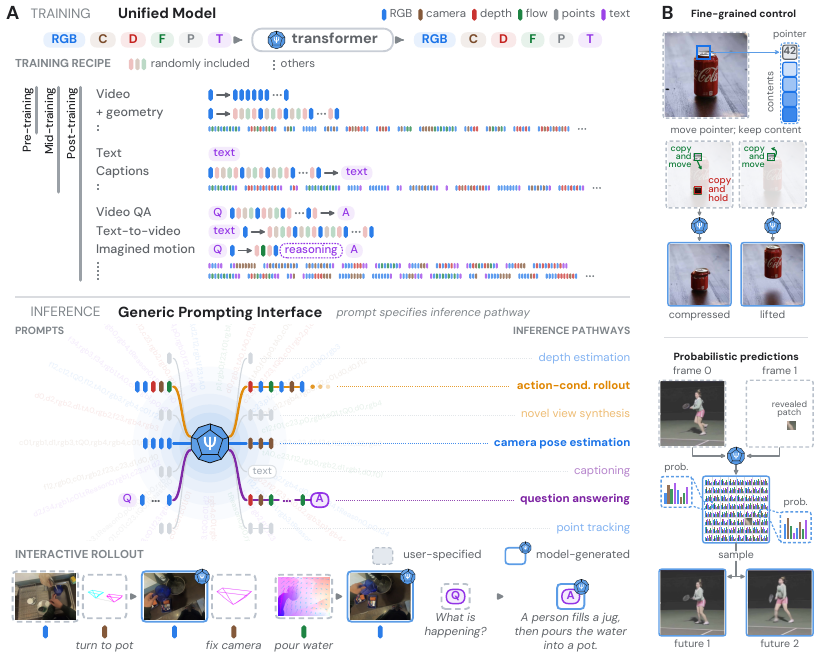}
\caption{\textbf{Architecture, training recipe, and prompting interface.}
\textbf{(A)}  $\Psi$-0.6 is a unified model across diverse modalities. Pretraining, mid-training, and post-training sample different sequence families; hollow tokens denote stochastically included sequence elements. 
Generic prompting interface: prompts specify the available information and requested inference pathway. $\Psi$-0.6 is interactive: it can generate a view based on a user-specified camera pose, then use that generated view as context for subsequent reasoning.
\textbf{(B)} Fine-grained control: each patch is represented by a pointer token and content tokens. Pointers specify modality and spatial and temporal locations, enabling targeted conditioning and scene edits. Probabilistic prediction: $\Psi$-0.6 represents a distribution over compatible continuations, allowing the same prompt to yield alternative predictions through sampling.}
\label{fig:architecture}
\vspace{-1em}
\end{figure}

$\Psi$-0.6 combines three conceptual components, extending a familiar LLM recipe beyond language: (i) a unified model provides a broad repertoire of visual, geometric, and linguistic capabilities; (ii) a generic prompting interface flexibly accesses and composes these capabilities; and (iii) reinforcement learning teaches the model to direct its own reasoning from task feedback. Here we explain how these components support \emph{capability unification}, \emph{compositional unification}, and \emph{optimization unification}, respectively.

\subsection{Unified Multimodal Model}
\label{sec:architecture}
\label{sec:principles}

\textbf{Architecture: one vocabulary, one transformer.}
Images, camera pose, depth, optical flow, point tracks, and text are serialized as tokens in a sequence processed by our 8B-parameter autoregressive transformer. Perception, description, and simulation thus become different conditional predictions.

\subsection{Generic Prompting Interface: Streams of Thought}
\label{sec:prompting}
\label{sec:sot}

A generic prompting interface makes the model's capabilities available as flexible reasoning operations. $\Psi$-0.6 reasons through \emph{streams of thought}: sequences interleaving images and text with explicit visual abstractions, including camera pose, depth, optical flow, and point tracks. These representations offer complementary \emph{projections of the world}, allowing each reasoning step to use a representation suited to the problem -- text, geometry, or a predicted scene.

\textbf{Any-to-any prompting: all tasks are token orderings.}
A prompt specifies what is given and what is requested. Different token choices and orderings express different inference pathways through the same model. Any modality combination can condition prediction, enabling richly controllable world modeling over spatial, temporal, and semantic aspects of a scene.
Let $\mathrm{RGB}_0$ denote an image, while $C_{0\to1}$ and $F_{0\to1}$ denote camera motion and optical flow. Then $\mathrm{RGB}_0,C_{0\to1}\rightarrow\mathrm{RGB}_1$ and $\mathrm{RGB}_0,\mathrm{RGB}_1\rightarrow C_{0\to1}$ express novel view synthesis and pose estimation; replacing $C$ with $F$ gives motion-conditioned generation and flow estimation. These are conditional predictions of the same model. A camera pose can therefore serve as an \emph{estimate} when inferred from images, a \emph{command} when supplied to generate a view, or an \emph{intermediate} when used to answer a spatial question. In this sense, the model acts as a ``\emph{multiverse generator}'' over a ``\emph{garden of forking paths}''~\citep{moire_language_2021}. Prompts control both {what is constrained} and {what is left for the model to infer}.

\textbf{Compositional reasoning: predictions become context.}
Streams of thought compose these conditional predictions into multistep reasoning processes, with the output of one prediction providing context for the next. For example, the model can generate a view from an image and a user-specified camera pose, then use the generated view as context for answering a spatial question (Fig.~\ref{fig:architecture}A). Visual and geometric predictions are therefore not only task outputs, but intermediate operations that connect perception, language, and imagination within one reasoning process. A prompt can prescribe these operations; their selection can also become part of what the model learns.

$\Psi$-0.6 builds on Probabilistic Structure Integration (PSI; $\Psi$)~\citep{kotar_world_2025}, inheriting two further properties that make its prompting interface unusually flexible: \emph{fine-grained control} over individual patches and \emph{probabilistic prediction}, which allows unspecified content to remain open to multiple plausible continuations. We kindly refer readers to Appendix Sec~\ref{sec:preliminaries_psi} for details.

\subsection{Reinforcement Learning: Adaptive Stream-of-Thought Reasoning}
\label{sec:adaptive_framework}

Beyond executing prescribed reasoning pathways, $\Psi$-0.6 learns \emph{representation-adaptive} streams of thought -- choosing \textit{which} representation to use and \textit{when} -- from task feedback. Within the shared token interface, these choices become policy actions, making the reasoning strategy itself a target of optimization.
We instantiate this approach through reinforcement learning and apply it to visual question answering. The policy selects visual representations, including depth, camera motion, and optical flow, according to the input and question, and receives reward based on downstream answer quality. Section \S\ref{sec:rl} specifies the learning formulation, training procedure, and evaluation.

\subsection{Training Recipe: Pretraining, Mid-training, and Post-training}
\label{sec:recipe}

Pretraining builds a broad base of predictive knowledge, mid-training integrates additional representations and data, and post-training teaches task-directed use of the resulting capabilities. In each supervised stage, we form different dataset groups (e.g., image captioning, video question answering), each with its own datasets and sequence notations, and allocate a training ratio to each. Training on these dataset groups is unified under a simple {next-token prediction} objective. Thus, sequence construction determines both what the model observes and what it learns to predict.

\textbf{Pretraining: visual world modeling and geometric integration.}
The model is pretrained on a large video dataset spanning natural scenes, human activity, and robot interaction. Training begins with RGB and camera conditioning and continues with depth and flow integrated into the sequence. Pure RGB sequences train appearance and motion prediction; interleaved sequences additionally train relationships between images and geometric structure. Keeping both in the mixture exposes the model to prediction with different amounts and types of conditioning information. This stage extends Probabilistic Structure Integration (PSI; $\Psi$)~\citep{kotar_world_2025}, part of a lineage of models demonstrating state-of-the-art visual understanding and world modeling across optical flow and point tracking~\citep{kim_taming_2025}, depth estimation and novel-view synthesis~\citep{lee_3d_2025}, object segmentation~\citep{venkatesh_discovering_2025}, physical object discovery and 3D manipulation~\citep{venkatesh_physical_2026}, and perceptual simulation~\citep{lee_perceptual_2026}. $\Psi$-0.6 extends these visual prediction and control capabilities to language-grounded reasoning in subsequent stages.

\textbf{Mid-training: vision-language alignment.}
The next step is learning language through text-only data from FineWeb-Edu and DataComp-LM (DCLM), followed by vision-language alignment by adding images and videos paired with captions, both global and temporally indexed.

\textbf{Post-training: useful, controllable, user-directed reasoning.}
Instruction tuning introduces image and video question answering and text-to-video control of camera and object motion. Post-training also teaches spatial reasoning (\S\ref{sec:mindcube}) and answering questions by counterfactually imagining object motion (\S\ref{sec:counterfactual}). Pretraining and mid-training sequences are retained to preserve the model's broad predictive capabilities. Beyond these next-token prediction stages, reinforcement learning trains adaptive streams of thought from task feedback, improving how the model selects intermediate representations for downstream reasoning (\S\ref{sec:rl}).

\begin{figure}[p]
\centering
\includegraphics[width=\textwidth]{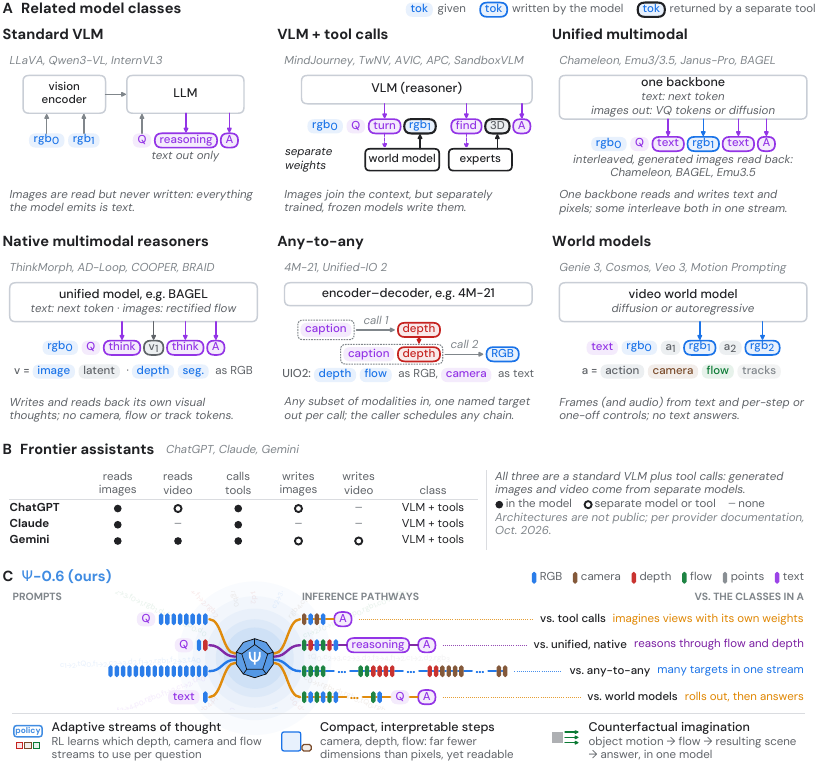}
\caption{\textbf{Comparison to related work.}
\textbf{(A)}~Six classes of related models.
\textbf{(B)}~Frontier assistants pair a VLM with separate tools.
\textbf{(C)}~$\Psi$-0.6 reads and writes all of these in one model; each pathway contrasts with a class in~(A).}
\label{fig:related_work}
\vspace{-1em}
\end{figure}

\section{Related Work}

\textbf{Unified multimodal models and reasoning.}
A growing line of work unifies increasingly broad sets of modalities and capabilities within a single model (Fig.~\ref{fig:related_work}).
Unified-IO, Unified-IO 2, and 4M-21 cast diverse vision, language, and geometric tasks into shared prediction interfaces \citep{lu_unified-io_2022,lu_unified-io_2023,bachmann_4m-21_2024}.
Chameleon and Emu3.5 extend this paradigm to interleaved image and text generation \citep{team_chameleon_2025,cui_emu35_2025}.
More recent work uses unified models for multimodal reasoning: ThinkMorph, AD-Loop, and BRAID interleave visual and textual reasoning steps \citep{gu_thinkmorph_2025,wu_synergizing_2026,hu_bridging_2026}, while COOPER incorporates depth and segmentation for spatial reasoning \citep{zhang_cooper_2025}.
These works demonstrate that shared multimodal representations can support not only diverse tasks, but also compositional reasoning.
$\Psi$-0.6 builds on these advances by integrating vision-language reasoning with physical world modeling through a shared token interface.
Sequences of text, images, camera pose, depth, optical flow, and point tracks form streams of computation, with generated representations recursively consumed and composed as intermediate reasoning steps.
This enables reasoning through explicit geometry and motion, including counterfactual simulation, while learning to select task-relevant representations from downstream feedback.

\textbf{Tool-augmented LLMs and VLMs.}
Another line of work enables compositional reasoning by allowing language or vision-language models to invoke external models, tools, or programs~\citep{schick_toolformer_2023,suris_vipergpt_2023,hu_visual_2024} (Fig.~\ref{fig:related_work}). In contrast, the intermediate operations in $\Psi$-0.6 are generated by the reasoner itself through the same shared token interface, rather than delegated to separately parameterized systems. This makes both the choice of intermediate representation and its contents part of the model's own computation, enabling end-to-end optimization of the reasoning process; we discuss the resulting advantages for co-adaptation, counterfactual generation, and task-dependent intermediate representations in \S\ref{sec:rl}.
\section{Capability and Compositional Unification: Spatial Intelligence}
\label{sec:mindcube}

\begin{figure}[t]
\centering
\includegraphics[width=\textwidth]{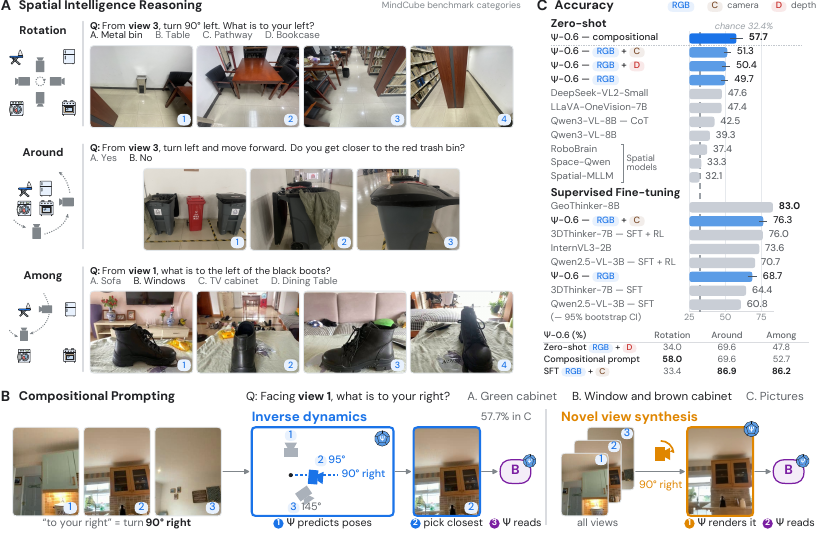}
\caption{\textbf{Spatial intelligence reasoning on \dataset{MindCube}.}
\textbf{(A)} Benchmark categories: Rotation, Around, and Among, adapted from \citet{wang_mindcube_2026}.
\textbf{(B)}~Two compositional-prompting routes: infer relative poses and select a real input view, or synthesize RGB at a target pose, then read an answer.
\textbf{(C)}~Accuracy for zero-shot and benchmark-trained systems; some trained comparisons use RL.}
\label{fig:mindcube}
\end{figure}


$\Psi$-0.6 represents our hypothesis for building unified VLMs with capable spatial intelligence. First, we start from a visual world model backbone trained to jointly model appearance, geometry, and motion~\citep{kotar_world_2025}, then extend it with language. Standard VLMs typically attach a visual encoder to a pretrained LLM, whose initial predictive knowledge comes from text~\citep{liu_improved_2024,alayrac_flamingo_2022}. Language pretraining does not directly supervise how appearance changes with viewpoint or object motion. We hypothesize that learning these relationships through visual and geometric prediction provides a stronger foundation for spatial reasoning, with language supplying an interface to these capabilities. Second, we ground reasoning in explicit geometric abstractions, whereas standard VLMs reason and answer primarily in text. A single transformer interleaves RGB frames, camera pose, depth, optical flow, point tracks, and text as a unified token sequence. These visual \emph{streams of thought} express spatial structure and motion in intermediate predictions that can be inspected to diagnose errors. Third, unifying world-model and vision-language capabilities yields compositional, end-to-end reasoning. Flexible token orderings express pose estimation, novel view synthesis, and question answering, allowing one task's output to condition the next. A spatial question can thus combine capabilities of the same model architecture -- estimate the camera motion the question describes, get to the view that reveals the answer, then read it.

\dataset{MindCube} tests spatial reasoning across multiple views of a scene~\citep{wang_mindcube_2026}. It is challenging in two ways: (i) its questions involve counterfactual camera motions (e.g., turn left) rather than answering from static images/video, and (ii) its images are separated by large angles (e.g., 90 degrees), which is unlike how most video models and VLMs are trained. 
\dataset{MindCube} is organized into three categories (Fig.~\ref{fig:mindcube}A), all requiring integrating observations across large viewpoint changes and reasoning in the reference frame specified by the question. 

\textbf{Zero-shot.}
Frozen $\Psi$-0.6 reaches $49.7$--$51.3\%$ across RGB-only, RGB-depth, and RGB-camera configurations (Fig.~\ref{fig:mindcube}C). $\Psi$-0.6 is state-of-the-art relative to open-source generalist VLMs (DeepSeek-VL2-Small, LLaVA-OneVision-7B, Qwen3-VL-8B), and spatial specialists (RoboBrain, Space-Qwen, Spatial-MLLM).

\textbf{Compositional Prompting.}
Beyond directly answering the question, $\Psi$-0.6 can combine its VLM capabilities with its vision and world-modeling capabilities (Figure~\ref{fig:mindcube}B). $\Psi$-0.6 uses \emph{inverse dynamics}: predict relative camera poses, select the real input view matching the question's requested turn, and read the answer from it. Alternatively, $\Psi$-0.6 uses \emph{novel view synthesis} (NVS): generate an RGB frame at the target camera pose and read the answer from this imagined view. Inverse dynamics requires the viewpoint to be one of the inputs, but NVS additionally allows a viewpoint absent from the inputs. Using inverse dynamics, $\Psi$-0.6 reaches $57.7\%$, compared to $51.3\%$ zero-shot.

\textbf{SFT.}
\dataset{MindCube} fine-tuning yields $68.7\%$ with RGB-only inputs and $76.3\%$ with RGB+camera inputs on the held-out Tiny test split (Fig.~\ref{fig:mindcube}C). The RGB-only result exceeds the displayed SFT-only Qwen2.5-VL and 3DThinker baselines. The RGB+camera result exceeds InternVL3's map-then-reason configuration and Qwen2.5-VL with SFT+RL, is comparable to 3DThinker with SFT+RL ($76.0\%$), and trails GeoThinker ($83.0\%$)~\citep{wang_mindcube_2026,chen_think_2026,li_geothinker_2026}.

\section{From Compositions to Counterfactuals}
\label{sec:counterfactual}

\begin{figure}[t]
\centering
\includegraphics[width=\textwidth]{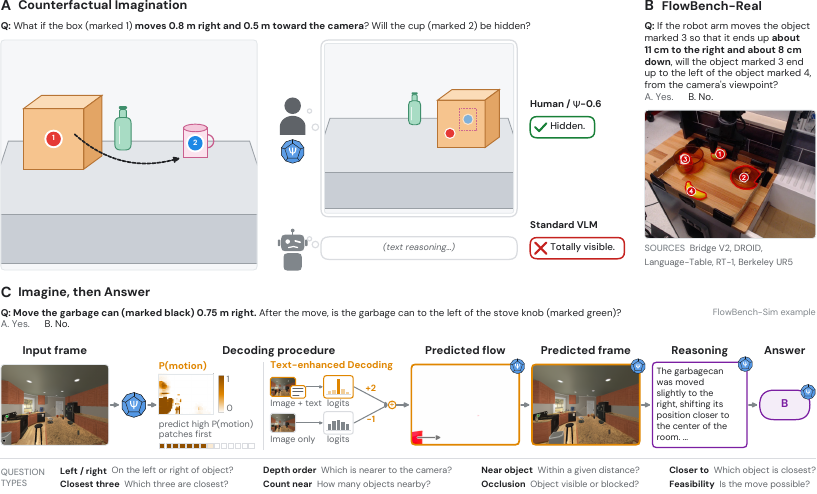}
\caption{\textbf{Counterfactual imagination of object motion.}
\textbf{(A)} Answering a counterfactual question requires imagining the outcome of a hypothetical motion. A person, like $\Psi$-0.6, imagines the box sliding in front of the cup and answers correctly, whereas a VLM that reasons only in text does not.
\textbf{(B)} A \dname-Real question on real robot data. $\Psi$-0.6 generalizes to \dname-Real after training on \dname-Sim (Tab.~\ref{tab:flowbench_real}).
\textbf{(C)} Stream of Thought on \dname-Sim: $\Psi$-0.6 predicts the optical flow of the instructed motion, decoding high-$P(\text{motion})$ patches first (pMotion) with text-enhanced CFG ($2\times$ image-and-text minus $1\times$ image-only logits), then the future frame, and reasons to its answer. Bottom: \dname{} question types.}
\label{fig:counterfactual_main}
\end{figure}

Beyond answering questions about static scenes, $\Psi$-0.6's WM capability lets it answer questions requiring \emph{counterfactual imagination}, reasoning about worlds that are not directly observed.
The model intervenes on a scene, predicts the consequence, and reads the answer off its own prediction. While \S\ref{sec:mindcube} moved the camera, here the intervention moves an object: the question states a displacement of a marked object and asks where things end up -- occluded or not, closer to which landmark, touching a wall, still feasible to push further, etc. This is the rung above association on the ladder of causation~\citep{pearl2009causality}, and it is exactly where existing open-source VLMs are near chance before fine-tuning.

\textbf{Benchmark.} \dataset{\dname}~\citep{liu_counterfactual_2026} evaluates visual world models on spatial questions involving counterfactual object motions. It comprises two splits: \dataset{\dname-Sim}, featuring scalable and accurate datasets in simulated indoor scenes~\citep{ai2thor,procthor}, and \dataset{\dname-Real}, augmenting five robot datasets (DROID, Bridge V2, Language-Table, RT-1, and Berkeley UR5) with counterfactual motion questions across diverse embodiments and environments (Appendix ~\ref{sec:flowbench}).

\begin{table}[h]
\centering\small
\caption{World modeling ability: expert models vs. $\Psi$-0.6.
``rEPE$_{\mathrm{move}}$'' is the relative endpoint error on moving pixels;
static pixels are excluded. ``Acc'' denotes what percentage of predictions have rEPE lower than a threshold. (Definition in Appendix.~\ref{suppl:def_flow})
``Blank'' stands for the percentage of the model predicting zero flow. 
``Dir'' is the 2-D angle between predicted and ground-truth flow on pixels where both move. 
``$\Delta$PSNR'' shows the improvement of the generated simulated image after the move on moving pixels compared with zero flow.
$n=622$.}
\label{tab:flow_epe_expert}
\resizebox{\textwidth}{!}{%
\begin{tabular}{lccccc|c|c}
\toprule
Method & Acc$\uparrow$ & rEPE$_{\mathrm{move}}$ $\downarrow$ & rEPE$_{\mathrm{move}}$ med $\downarrow$ & Blank $\downarrow$& Dir ($^\circ$) $\downarrow$& $\Delta$PSNR $\uparrow$& Oracle Input \\
\midrule
Zero flow & 0.0\%&1.000 & 1.000 & 100.0\% & -- &0.00& -- \\
\midrule
Modular {\scriptsize(Grounding DINO + SAM2 + Depth Pro)} & \underline{50.8\%}& \underline{0.552} & \underline{0.358} & 15.0\% & \textbf{1.4} &-- &Camera intrinsics\\
\midrule
FOFPred~\citep{ran_fofpred_2026} & 2.4\%& 0.915 & 0.958 & \textbf{0.0\%} & 25.9 & --&-- \\
Qwen-Image-Edit~\citep{wu_qwenimage_2025} & 7.9\%& 2.154 & 1.014 & 2.9\% & 64.8 & -1.63&GT frame-0 mask \\
IP2P~\citep{brooks2023instructpix2pix} {\scriptsize(plain MSE)} & 10.5\%& 2.659 & 1.221 & 2.9\% & 71.0 & -3.35&GT frame-0 mask \\
IP2P~\citep{brooks2023instructpix2pix} {\scriptsize(SoM target)} & 13.7\%& 2.938 & 1.122 & 3.1\% & 64.4 &-0.96 &GT frame-0 mask \\
IP2P~\citep{brooks2023instructpix2pix} {\scriptsize(mask-weighted)} & 16.4\%& 2.655 & 1.000 & 2.7\% & 57.3 & -4.00&GT frame-0 mask \\
\midrule
$\Psi$-0.6 Default seq 32 & 31.8\% & 0.635 & 0.639 & 27.5\% & 14.1 & \textbf{+1.60}	& --\\
$\Psi$-0.6 + pMotion + CFG2.0 & \textbf{56.4\%}& \textbf{0.454} & \textbf{0.356} & \underline{0.3\%} & \underline{8.1} &\underline{+0.94}&-- \\
\bottomrule
\end{tabular}%
}
\end{table}

\textbf{World modeling: $\Psi$-0.6 outperforms state-of-the-art expert models.}
Text- and action-conditioned future prediction is one of the pressing abilities needed in world modeling, with promising applications like planning in robotics. 
To improve text-to-flow predictions, $\Psi$-0.6 employs two powerful decoding techniques. 
Critically, these are enabled by $\Psi$-0.6's fine-grained control and probabilistic predictions, attributes that are absent from many other world models. 
First, before generating flow, $\Psi$-0.6 computes the probability of motion (``\emph{pMotion}'') at every patch with a single feedforward pass. Then, $\Psi$-0.6 generates flow predictions sequentially from patches most likely to move to least likely. The intuition behind this is: \yifan{in \dataset{\dname}, motion typically involves a single, often small object, so predicting flow for the patches most likely to move first helps the model focus on the relevant region.} 
Second, our model uses a form of classifier-free guidance (CFG)~\citep{ho2022cfg}: at every generated token, compare the flow prediction logits with and without conditioning on the text instruction, then amplify the difference -- this amplifies the effect of the text instruction on motions. 
Results for ablating these two techniques are in Appendix \ref{sec:flow_pred_ablation}.

We compare our models with various baseline models: FOFPred~\citep{ran_fofpred_2026}, InstructPix2Pix~\citep{brooks2023instructpix2pix}, Qwen-Image-Edit~\citep{wu_qwenimage_2025}, and a Modular design with state-of-the-art experts~\citep{liu_grounding_2024,ravi_sam_2025,bochkovskiy2025depth} (Appendix ~\ref{sec:baselines}).
Against baselines, $\Psi$-0.6 establishes the strongest overall flow prediction without privileged inputs. Its best configuration reaches $56.4\%$ accuracy, compared with $16.4\%$ for the strongest image-editing baseline. It also reduces rEPE$_{\mathrm{move}}$ to $0.454$, versus $0.915$ for FOFPred. Even the modular geometric pipeline reaches only $50.8\%$ accuracy and $0.552$ rEPE$_{\mathrm{move}}$, despite using specialized models for each task and the oracle input. Finally, every evaluated $\Psi$-0.6 configuration improves next-frame prediction fidelity over the zero-motion reference ($+0.25$ to $+1.60$ dB, for the moving region), whereas the image editors decrease it ($-0.96$ to $-4.00$ dB). Together, these results show that $\Psi$-0.6 predicts both where an object moves and how its appearance should change more reliably than task-specific generative experts, while requiring no oracle inputs.




\begin{table*}[t]
\centering
\small
\setlength{\tabcolsep}{4pt}
\caption{Accuracy on \dname-Sim (\%). Closest-3 requires the exact set of three. Cat.\ avg is the unweighted mean over the 8 categories.
}
\label{tab:cit}
\resizebox{\textwidth}{!}{%
\begin{tabular}{lcccccccc|cc}
\toprule
Model & Closer & Closest\_3 & Contact & Count & Feasibility & L/R & Near & Occl. & Avg & Cat.\ Avg \\
\midrule
Chance & 25.0 & 5.0 & 50.0 & 25.0 & 50.0 & 50.0 & 50.0 & 50.0 & 35.8 & 38.1 \\
\midrule
Qwen3-VL-8B~\citep{Qwen3-VL} & 47.2 & 11.9 & 52.6 & 24.5 & 56.6 & 77.3 & 58.4 & 55.0 & 46.9 & 47.9 \\
Qwen3.5-9B~\citep{Qwen3.5} & 53.5 & 10.7 & 52.6 & 23.3 & 52.8 & 68.4 & 53.3 & 51.7 & 44.3 & 45.8 \\
Qwen3-VL-32B~\citep{Qwen3-VL} & 41.8 & 17.0 & 38.6 & 20.9 & 67.6 & 81.9 & 55.3 & 62.5 & 48.3 & 48.2 \\
Qwen3-VL-8B \dname-FT & \textbf{89.0} & \textbf{36.3} & \textbf{93.0} & \textbf{57.3} & \textbf{88.4} & \textbf{98.6} & \textbf{88.0} & \underline{72.5} & \textbf{76.3} & \textbf{77.9} \\
\midrule
$\Psi$-0.6 Base & 23.4 & 0.0 & 24.0 & 12.1 & 9.4 & 45.7 & 50.4 & 31.5 & 24.5 & 24.6 \\
$\Psi$-0.6 FT & \underline{85.1} & \underline{33.9} & \underline{73.7} & \underline{48.2} & \underline{85.9} & \underline{97.4} & \underline{87.2} & \textbf{77.5} & \underline{73.1} & \underline{73.6} \\
\bottomrule
\end{tabular}%
}
\end{table*}

\begin{table*}[t]
\centering
\small
\setlength{\tabcolsep}{4pt}
\caption{Accuracy on \dname-Real (\%). Note: No \dname-Real examples are used in training, so this table evaluates the generalization ability of models.
Closest-3 requires the exact set of three. Cat.\ avg is the unweighted mean over the 7 categories.
}
\label{tab:flowbench_real}
\resizebox{\textwidth}{!}{
\begin{tabular}{lccccccc|cc}
\toprule
Model & L/R & Near & Closer & Depth & Closest-3 & Count & Occl. & Avg & Cat.\ Avg \\
\midrule
Chance &  50.0 & 50.0 & 26.7 & 50.0 & 7.0 & 25.1 & \underline{50.0} & 37.6 & 37.0 \\
\midrule
Qwen3-VL-8B~\citep{Qwen3-VL}  &  52.2 & 48.9 & 42.3 & \underline{62.1} & 7.9 & \underline{38.2} & 37.2 & 42.7 & 41.2 \\
Qwen3.5-9B~\citep{Qwen3.5} & 52.2 & 56.1 & 40.6 & 50.0 & 11.4 & 30.0 & \textbf{50.7} & 42.7 & 41.6 \\
Qwen3-VL-32B~\citep{Qwen3-VL} & \underline{58.9} & 49.9 & 40.7 & 50.0 & 19.0 & \textbf{43.5} & 39.2 & 44.4 & 43.0 \\
Qwen-8B \dname-FT & \textbf{70.3} & \textbf{80.0} & \underline{61.4} & \textbf{62.8} & \textbf{58.0} & 29.5 & 31.9 & \textbf{59.6} & \textbf{56.3} \\
\midrule
$\Psi$-0.6 Base  & 50.3 & 51.3 & 23.3 & 50.0 & 7.4 & 26.7 & \underline{50.0} & 37.5 & 37.0 \\
$\Psi$-0.6 \dname-FT   & 57.5 & \underline{74.6} & \textbf{63.0} & 41.1 & \underline{29.6} & 26.7 & \underline{50.0} & \underline{51.4} & \underline{48.9} \\
\bottomrule
\end{tabular}
}
\vspace{2em}
\end{table*}

\textbf{Strong VQA abilities of $\Psi$-0.6.}
Fine-tuning on \dataset{\dname-Sim} greatly improves $\Psi$-0.6's capabilities: accuracy rises from 24.5\% to 73.1\% overall and from 24.6\% to 73.6\% category-averaged, improving in all eight categories and roughly doubling the 35.8\% chance level (Table~\ref{tab:cit}). This places $\Psi$-0.6 close to Qwen3-VL-8B fine-tuned on the same data, a dedicated VLM designed for Visual Question Answering.
$\Psi$-0.6 is the only model we evaluate that is at once the strongest flow predictor and a competitive counterfactual question answerer.

To evaluate transfer, we test the models on \dataset{\dname-Real} without using any of its examples during training (Table~\ref{tab:flowbench_real}). $\Psi$-0.6 reaches 51.4\% overall and 48.9\% category-averaged accuracy, improving by 13.9 and 11.9 percentage points over the base model. Transfer is clearest in the categories that depend on metric geometry rather than on language priors: ``Closer'' rises from 23.3\% to 63.0\%, and ``Near'' reaches 74.6\%, both ahead of every open-source model we evaluate without fine-tuning.
This shows that supervised fine-tuning of $\Psi$-0.6 on \dataset{\dname} enables \emph{sim-to-real} transfer for its spatial question-answering capabilities to unseen data. Viewed alongside its world-modeling capabilities, these results support the use of $\Psi$-0.6 as a unified framework for world modeling and VQA.
\section{Optimization Unification: RL for Adaptive Streams of Thought}
\label{sec:rl}

\begin{quote}
``My experience is what I agree to attend to.'' \\
\hfill --- William James, \textit{The Principles of Psychology}
\end{quote}

\begin{figure}[t]
\centering
\includegraphics[width=\textwidth]{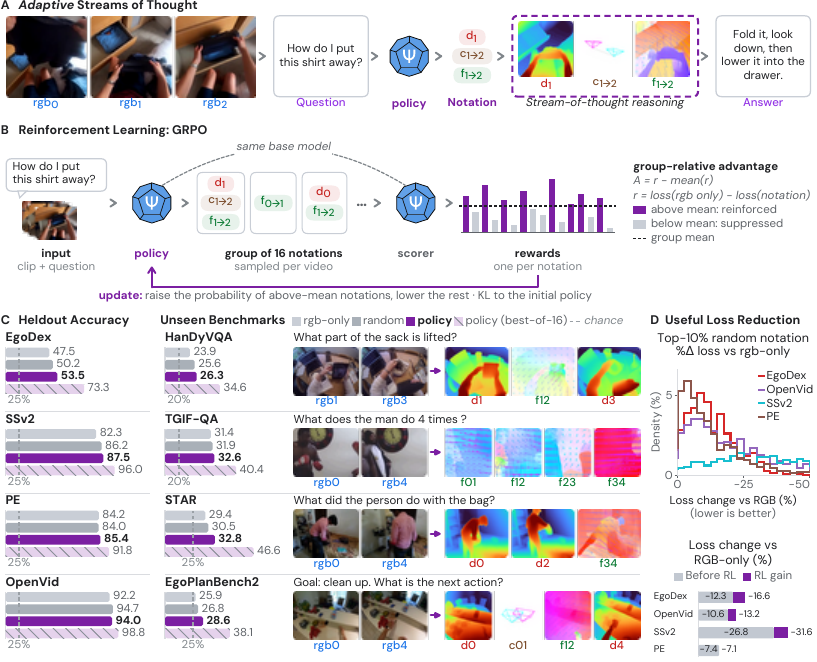}
\caption{\textbf{Learning adaptive stream-of-thought reasoning.}
\textbf{(A)}~Given a video and question, $\Psi$-0.6 functions as both a policy -- emitting a notation naming the stream of thought to use -- and as a VLM, answering with those streams in context.
\textbf{(B)}~The GRPO RL loop: collect notations with the current policy, score each by probability of ground-truth answer relative to RGB-only, and train on the beneficial notations. 
\textbf{(C)} The learned $\Psi$-0.6 policy improves video question-answering accuracy, with gains generalizing to unseen benchmarks.
\textbf{(D)} These accuracy gains are a result of loss reductions on the ground truth.
}
\label{fig:select_modality}
\end{figure}

Whereas $\Psi$-0.6 used prescribed reasoning pathways in the previous sections, here we show that it can learn streams of thought that are \emph{self-generated}, \emph{representation-adaptive}, and unfold over time. As reasoning proceeds, the model learns not only what intermediate content to generate, but which representation to invoke and when. Unification makes both the choice and generation of these intermediates into actions of the same policy, enabling reinforcement learning on task feedback to optimize the reasoning process end to end (\emph{optimization unification}).

\begin{itemize}
[leftmargin=*]
\item \textbf{Self-generated intermediates.} Unlike tool-augmented LLMs and VLMs that invoke external models or programs~\citep{schick_toolformer_2023,suris_vipergpt_2023,hu_visual_2024}, $\Psi$-0.6 generates its intermediates within the reasoner itself. This enables: (i) \emph{co-adaptation} of the reasoner and its intermediate predictions through gradient-based credit assignment, (ii) generation of states unavailable through existing tool calls, including future or counterfactual states, and (iii) task-dependent intermediates whose appropriate form may itself be ambiguous. For example, a bottle and its cap are one segment for object recognition, yet separate segments for robotic manipulation.

\item \textbf{Representation-adaptive computation.} $\Psi$-0.6 uses {representation}-adaptive \emph{test-time compute} through explicit geometric projections of the world. In text chain-of-thought, the representational medium is fixed -- language; in $\Psi$-0.6, the medium itself becomes part of the computation.

\item \textbf{Unfolds over time.} Different representations can be called into computation as the situation evolves: depth for spatial structure, optical flow for motion, camera pose across viewpoints, or a predicted scene to imagine what may happen next. This is particularly relevant to embodied agents and robotics, where perception and action unfold continuously in the world.
\end{itemize}

\textbf{Learning to Select Streams of Thought}
\label{sec:policy}
Although our framework is generic, here we show a proof-of-concept specifically for visual question-answering. Given a video $v$ and question $q$, the policy emits a notation $z$ selecting from depth, camera-motion, and optical-flow streams (Fig.~\ref{fig:select_modality}A). For example, $d_0,c_{2\to3},f_{2\to3}$ requests depth at the first frame and camera motion and flow between two later frames. This turns $\Psi$-0.6's generic prompting interface into a policy over reasoning representations. 
At a high level, reinforcement learning alternates between two phases (Fig.~\ref{fig:select_modality}B):
\begin{itemize}
[leftmargin=*]
\item \textbf{Collect data with the policy.} For each question, sample notations $z$ from the current policy (initialized with a uniform distribution), and score each by how much it lowers the ground truth's loss. We build a dataset using videos from EgoDex~\citep{hoque_egodex_2026}, Something-Something~v2 (SSv2)~\citep{goyal_something_2017}, PE~\citep{bolya_perception_2025}, and OpenVid-1M~\citep{nan_openvid-1m_2025}. Since these datasets provide captions rather than questions, we convert their captions into question-answer pairs (Appendix~\ref{sec:pwpn_data}).
\item \textbf{Train the policy on the collected data.} Initially, we use filtered behavior cloning (rejection-sampling fine-tuning) on the sampled notations that beat RGB-only to teach the model to emit notations; subsequently, we use on-policy GRPO (Group Relative Policy Optimization) to update the policy's LoRA parameters with rewards relative to the other samples in the group, refining which streams it selects.
\end{itemize}


\textbf{Learned policy generalizes accuracy improvements to unseen benchmarks.}
The RL-trained policy improves accuracy on held-out examples from the training distributions (a mean relative gain of 5.6\% over video-only question answering), with the largest gains on EgoDex, the most challenging dataset. Encouragingly, this suggests that adaptive reasoning may provide greater benefits on harder problems, where additional intermediate computation is most useful. Critically, these gains transfer to four completely unseen benchmarks (a mean relative gain of 9.0\%), providing evidence that the learned policy generalizes beyond its training data. Another encouraging signal is the best-of-$k$ performance is substantially higher than single-rollout (Fig.~\ref{fig:select_modality}C), indicating further headroom for policy optimization. The selected streams are diverse and, in some cases, interpretable; for example, repeated-action questions can elicit sequences of optical-flow predictions, object interactions can combine depth and flow, and observer-motion questions can combine depth with camera motion.

\section{Discussion}
\label{sec:discussion}
We introduce $\Psi$-0.6, a unified vision-language world model that turns visual, geometric, linguistic, and world-model predictions into composable reasoning operations. Through capability, compositional, and optimization unification, a single model can perceive, imagine, reason, and learn which representations to invoke as a problem unfolds. Streams of thought support stronger spatial reasoning, counterfactual imagination, and representation-adaptive reasoning learned from task feedback. Importantly, the medium of reasoning need not be fixed: models can learn not only \emph{what} to predict, but \emph{which representation} to reason through and \emph{when}. This opens a path toward agents whose reasoning continually reorganizes itself across representations as they learn from embodied experience.

More broadly, reinforcement learning over streams of thought could support a remarkably general recipe for representation-adaptive reasoning across diverse domains and modalities. Within a shared predictive interface, observations in any modality can supply feedback for learning how to reason through others. For example, given a video, a stream of thought through depth, optical flow, or camera motion could be rewarded by how well it supports predictions of future visual input. This serves as a label-free, task-independent source of reward without human annotations or separate RL environments. Once perception and prediction are part of the reasoner, observations of the world can provide supervision for not only {what} an embodied agent predicts, but {which representation} of the physical world to reason through before acting. Extending $\Psi$-0.6 with actions and proprioception would allow perception, physical reasoning, imagination, and control to become composable operations of a single agent.

\subsection{Implications for physical reasoning and learning}

\paragraph{Benefits of geometric representations in stream-of-thought reasoning.}
First, geometric representations provide an explicit \emph{decomposition of the problem}. One might ask why such intermediates should be useful when all the information is contained in the original video itself. But the same is true in mathematical reasoning: all the information is contained in the problem statement, yet explicitly producing and using intermediate chains of thought \citep{wei_chain_of_thought_2022} can decompose a difficult computation into simpler steps, improving a model's final output. Geometric representations can play an analogous role for visual reasoning, exposing depth, motion, correspondence, and camera transformation rather than requiring them to be repeatedly inferred from pixels. Empirically, we find that using these intermediates improves both understanding, such as video question-answering, and fine-grained prediction and control, such as reasoning through optical flow or camera motion.

Second, geometric representations can focus learning on the crucial aspects of the physical world. These are {lower-dimensional} projections that deliberately \emph{abstract} away incidental visual detail while preserving {factors of the environment} -- such as geometry, motion, correspondence, and viewpoint -- that might generalize beyond the particular objects, scenes, or environments encountered during training. Predicting several such representations from the same experience can also provide a form of ``data augmentation'': a single video supplies multiple aligned, structured views of the same underlying physical event. Together, these may improve \emph{sample efficiency}, particularly in regimes where both the quantity and diversity of data are limited \citep{aw_zero-shot_2026}, such as robotics, biology, or the experimental sciences.

Third, these intermediates combine the twin advantages of \emph{interpretability} and \emph{compactness}. ``Thinking with images/video'' reasons by generating or editing {pixels}, making intermediate states directly interpretable~\citep{wiedemer_video_2025,tong_thinking_2026,li_imagine_2025,chern_thinking_2025,bian_optical_2026}, whereas {latent} reasoning compresses intermediate computation into continuous representations that are difficult to inspect directly~\citep{wang_monet_2025,li_latent_2025,yang_machine_2025,zhang_latent_2025,wu_lavit_2026}. $\Psi$-0.6 occupies a middle ground by {externalizing} compact intermediates as physically meaningful tokens: their semantics are grounded in recognizable properties of the physical world, but they need not reproduce every visual detail to expose the structure relevant to a computation. 

\subsection{Limitations}

\paragraph{Scale.}
Our results establish the conceptual and functional consequences of unifying vision, language, and world modeling, rather than exhaust the performance achievable by this framework. $\Psi$-0.6 was developed at substantially smaller scale in training data, compute, research, and engineering effort than contemporary frontier multi-modal systems. We expect scaling these resources will strengthen all of $\Psi$-0.6's capabilities; our preliminary experiments already suggest continued gains with scale, motivating further study.

\paragraph{Learning synergy and specialization.}
$\Psi$-0.6 advances the broader trend toward unified models by bringing vision, language, and world modeling into a shared reasoning process. Unification can promote positive transfer through shared knowledge~\citep{longpre_flan_2023}, but can also introduce interference when capabilities compete for limited capacity or require conflicting computations~\citep{conneau_unsupervised_2020,shi_realunify_2026,wu_uncovering_2026}. $\Psi$-0.6 demonstrates {functional unification} and inference-time gains from composing capabilities, but we do not isolate positive and negative transfer among those capabilities during training. A natural next step is a {sparse mixture-of-experts} architecture, which could retain the advantages of unification while managing the cost of a large model~\citep{shazeer_outrageously_2017,mustafa_multimodal_2022}. By activating only a subset of experts per token, it allows specialization and greater total capacity without a proportional increase in inference computation.

\subsection{Key open problems}

\paragraph{From visual proxy actions to embodied control.}
Our next step is extending the framework from reasoning about actions to acting in the physical world. In $\Psi$-0.6, optical flow and point tracks already function as \emph{proxy actions}: they specify how parts of a visual scene should move, but do not directly actuate the world. One route toward robotics would be to train an inverse-kinematics (IK) model conditioned on flow or point predictions, analogous to RGB-based IK models~\citep{goswami2025robopepp,liu2025robotag,ban2024real}. More generally, robot joint commands, end-effector poses, gripper states, and other control signals could become additional token modalities. Policy prediction and world simulation could then become different queries to the same model, extending the unified interface from perception and imagination to physical control.

Combining such action tokens with \emph{adaptive streams of thought} would enable a richer form of embodied reasoning. Rather than mapping observations directly to commands, the model could learn when to inspect geometry, predict motion, imagine the consequences of an action, evaluate progress, seek another observation, or act. Our current adaptive-stream experiments optimize downstream task performance on visual question answering rather than physical control. In an embodied agent, reward could jointly optimize {what action to take}, {what reasoning to perform before acting}, and {how much computation that reasoning should consume}. This would extend our adaptive streams of thought from choosing how to reason about the world to choosing how to reason {and act} within it.

\paragraph{Generic verification for the physical world.}
Physical reasoning may provide an unusually rich source of supervision because different representations of the same world must satisfy shared constraints. Depth, camera motion, optical flow, point tracks, and appearance should agree across viewpoints and time, while forward and inverse predictions create additional cycles that can be checked. For example, a camera motion inferred from two views should generate a compatible second view when supplied back to the model. Recent work shows that cross-modal cycle consistency can itself provide a dense, label-free reward for multimodal reasoning~\citep{zhang_rc2_2026}. Such signals suggest a route toward generic \emph{physical verifiers}: mechanisms that score intermediate reasoning through consistency with the structure of the world rather than requiring human labels for every trajectory.

\paragraph{Autodiscovery.}
Verification could, in turn, support an automatic loop for the discovery of reasoning intermediates. $\Psi$-0.6 currently reasons through deliberately chosen abstractions whose physical meaning is known in advance. Adaptive streams of thought learn \emph{which} of these representations to use and \emph{when}; future systems could instead ask \emph{which intermediate representations should exist at all}. A model could propose compact intermediate variables, evaluate whether reasoning through them improves predictive consistency or downstream task performance \citep{stojanov_self-supervised_2025}, and retain useful abstractions within its own reasoning interface. Work on latent-action world models already shows that useful transition representations can emerge from unlabeled video rather than being specified a priori~\citep{gao_adaworld_2025,ren_videoworld_2025}. Extending this idea from action representations to reasoning representations suggests a closed loop: propose an abstraction, evaluate the consequences of using it, and reincorporate successful abstractions into future computation. In this view, not only the reasoning policy, but the vocabulary through which a model reasons, can itself become an object of learning.

\section*{Acknowledgments}
We are very grateful to Yash Shah, Kristine Zheng, Cliona O'Doherty, Seojin Lee, Arjun Chandra, Greyson Brothers, Khaled Jedoui, and Ishaan Gupta for helpful feedback on the manuscript.

\paragraph*{Funding:}
This work was supported by the following awards. D.L.K.Y.: Simons Foundation grant 543061, National Science Foundation CAREER grant 1844724, National Science Foundation Grant NCS-FR 2123963, Office of Naval Research grant S5122, ONR MURI 00010802, ONR MURI S5847, and ONR MURI 1141386 - 493027.
We also thank Stanford HAI, Stanford Data Sciences, Stanford Marlowe team, and the
Google TPU Research Cloud team for computing support.



\paragraph*{Data and materials availability:}
Our model and code will be released at \url{https://huggingface.co/StanfordNeuroAILab} and \url{https://github.com/neuroailab}.




\bibliographystyle{unsrtnat_etal}
\bibliography{psi06,psi_vlm_imported_khai}

\appendix
\clearpage
\newpage

\section{Supplementary Results}

\subsection{Spatial Intelligence Reasoning: MindCube}

An interesting observation is that different spatial tasks expose different bottlenecks. \dataset{MindCube} separates in-place camera turns (Rotation), movement around a scene (Around), and relative object positions across views (Among; Fig.~\ref{fig:mindcube}A). Direct RGB+depth prompting is strongest on Around and near chance on Rotation. Structured prompting improves the displayed Rotation and Among results most. The SFT configuration is strong on Around and Among but remains near chance on Rotation -- we hypothesize that this is because the Rotation training data contains very limited scene diversity, making it challenging for learning to generalize to the test set.

\textbf{Inference protocol for the fine-tuned models.}
All fine-tuned $\Psi$-0.6 results on \dataset{MindCube} (Fig.~\ref{fig:mindcube}C) use \emph{direct answering}: the model emits the answer immediately after the inputs, without generating a text reasoning trace. We fine-tune on the 10k \dataset{MindCube} training questions -- which are disjoint from the 1{,}050-question Tiny split used for evaluation -- mixed with the model's general training data. Each training sequence places the question and its answer options first, then the input views (optionally with depth and/or relative camera poses), and ends with the answer.

\subsection{Details of FlowBench}
\label{sec:flowbench}
\dataset{\dname}~\citep{liu_counterfactual_2026} is a robotics and computer vision benchmark that evaluates the capability of visual world models to answer questions based on specified counterfactual motions. \dataset{\dname} contains diverse categories of questions, such as ``occlusion'' (``if we move the table 0.50 meters to the right, will it occlude the view of the chair?'') and ``closer to'' (``if we move the eraser 5cm to the left, which three objects would it be closest to?''). \dataset{\dname} contains two main splits. First, \dataset{\dname-Sim} is built in a simulator~\citep{ai2thor,procthor} and contains diverse indoor scenes with full rooms depicted, and involves moving objects with various scales from chairs to potatoes. 
Second, \dataset{\dname-Real} was built by adding counterfactual motion questions to five robot datasets: DROID, Bridge V2, Language-Table, RT-1, and Berkeley UR5. It spans five distinct robot embodiments, with scenes ranging from laboratory tabletops to toy kitchens and kitchen counters, and moved objects ranging from colored blocks and toy food to cups, bowls, bottles, utensils, and cloths.


\subsection{Details for baselines in flow prediction}
\label{sec:baselines}
\textbf{Baselines.} We compare our model against expert models that can map one RGB frame plus a language instruction to a dense displacement field. Each is fine-tuned on the same \dname training set used for training $\Psi$-0.6, and evaluated on the identical benchmark. We first report a zero-flow predictor: under a static camera only 0.6\% of pixels move, so predicting nothing is already a fine approximation, and any method whose wrong vectors exceed the true magnitude scores worse. 
FOFPred~\citep{ran_fofpred_2026} is the closest published match, image plus action phrase to HSV-encoded flow, fine-tuned with LoRA~\citep{hu_lora_2021} after raising its magnitude normaliser so our displacements are not clipped. As instruction editors we fine-tune InstructPix2Pix~\citep{brooks2023instructpix2pix} in three variants differing only in supervision target (badge-free render, Set-of-Mark render, and a mask-weighted loss on the object's before/after footprints), and Qwen-Image-Edit~\citep{wu_qwenimage_2025} with 4-bit QLoRA~\cite{dettmers2023qlora}. The mask weighting targets an identity shortcut: rgb0 and rgb1 differ in under 1\% of pixels on average, so a plain reconstruction loss is minimised by copying the input. Finally, a non-generative pipeline — the only baseline using the instruction's metric distance — grounds the referring expression with Grounding DINO~\citep{liu_grounding_2024} and SAM 2~\citep{ravi_sam_2025}, predicts metric depth with Depth Pro~\citep{bochkovskiy2025depth}, and applies the displacement as a 3-D rigid translation.

\textbf{Read-out and oracles.} Editors emit images, not flow, and correspondence models cannot convert them here: with an identical camera the correct warp is the identity, so given the true rgb1, SEA-RAFT~\citep{wang2024sea}, RoMa~\citep{edstedt_roma_2024} and CoTracker3~\citep{karaev2025cotracker3} recover under 25\% of the true magnitude. We instead localise the object by frame-difference proposals scored with DINOv2 descriptors~\citep{oquab2023dinov2} and refined by SAM 2. This paints flow on the ground-truth frame-0 mask, so editors are given that mask while PSI and FOFPred must also decide which pixels move — the groups are not directly comparable and PSI's margin is understated. \textbf{Notably, $\Psi$-0.6 does not need any of these oracles to perform well on \dname}, as it is able to read the answer off its own prediction.

\subsection{Definition of metrics for flow prediction} 
\label{suppl:def_flow}

We specify the definition of rEPE and Acc for
Table~\ref{tab:flow_epe_expert}. Let $f \in \mathbb{R}^{H \times W \times 2}$ be the
ground-truth flow from $I_0$ to $I_1$ and $\hat{f}$ the prediction, bilinearly resampled
to the ground-truth resolution $H \times W = 600 \times 800$ of FlowBench when a model decodes at a
different size. We write $e_p = \lVert \hat{f}_p - f_p \rVert_2$ for the endpoint error at
pixel $p$ and $\mathcal{M} = \{\, p : \lVert f_p \rVert_2 > 1\,\text{px} \,\}$ for the set
of ground-truth moving pixels. Because each counterfactual displaces a single object in
front of a static camera, $\mathcal{M}$ covers only ${\sim}0.6\%$ of the frame while the
displacements inside it are large (mean $100.2$\,px, median $83$\,px). An endpoint error
averaged over the whole frame is therefore dominated by static background and is nearly
identical for every method, and an absolute error restricted to $\mathcal{M}$ conflates a
small mistake on a large motion with a large mistake on a small one. We thus report
a \emph{relative} error on the moving pixels as the
primary quantity.

\textbf{rEPE.} Each pixel's error is normalised by that pixel's own motion distance before
averaging,
\begin{equation}
\mathrm{rEPE}_{\mathrm{move}}(\hat{f}, f)
  = \frac{1}{|\mathcal{M}|} \sum_{p \in \mathcal{M}}
    \frac{\lVert \hat{f}_p - f_p \rVert_2}{\lVert f_p \rVert_2}.
\label{eq:repe}
\end{equation}
The normalisation is per pixel rather than per sample or per dataset: a prediction is
charged by the \emph{fraction} of the true displacement it gets wrong. Static pixels are
excluded, since a relative error against zero motion is undefined. This fixes the scale of
the column: a zero-flow prediction ($\hat{f} \equiv 0$) scores exactly $1.000$ on every
sample, so any entry above $1$ is worse than predicting no motion at all, and $0.5$ means
the residual error is half the motion it was meant to reproduce. The table reports the mean
of Eq.~\ref{eq:repe} over the $N = 622$ benchmark samples; the ``med'' column is its median
over samples, which is robust to the few cases where a method moves the wrong object and
incurs an arbitrarily large error.

\textbf{Acc.} We additionally report the fraction of samples whose relative error falls
below a tolerance $t$,
\begin{equation}
\mathrm{Acc}(t) = \frac{1}{N} \sum_{i=1}^{N}
  \mathbbm{1}\!\left[\, \mathrm{rEPE}_{\mathrm{move}}\big(\hat{f}^{(i)}, f^{(i)}\big) \le t \,\right],
  \qquad t = 0.4 .
\label{eq:acc}
\end{equation}
A prediction counts as correct when its endpoint error is, averaged over the moved object,
within $40\%$ of the true displacement. The tolerance is relative rather than a fixed pixel
radius, so it scales with the motion: a $100$\,px displacement is allowed $40$\,px of slack,
a $30$\,px displacement only $12$\,px. By construction a blank prediction scores
$\mathrm{rEPE}_{\mathrm{move}} = 1$ and is never counted as correct, so a model cannot
inflate Acc by abstaining. We use $t = 0.4$ for every row of
Table~\ref{tab:flow_epe_expert}.


\subsection{VQA on \dataset{\dname}.}

Fine-tuning on \dataset{\dname} substantially improves the VQA performance of $\Psi$-0.6 across all eight categories (Table~\ref{tab:cit}). Overall accuracy increases from 24.5\% to 73.1\%. The fine-tuned model outperforms all evaluated off-the-shelf Qwen models on both aggregate metrics and achieves the second-highest overall accuracy, trailing Qwen3-VL-8B fine-tuned on the same dataset by 3.2 points. At the category level, $\Psi$-0.6 stays within 4 points of that model on five of the eight categories and is ahead on Occlusion (77.5\% vs.\ 72.5\%); the remaining gap is concentrated in Contact (73.7\% vs.\ 93.0\%) and Count (48.2\% vs.\ 57.3\%). These results demonstrate strong abilities in $\Psi$-0.6's spatial question answering, showcasing $\Psi$-0.6's capability to serve as a VLM.

\subsection{Flow prediction ablation}
\label{sec:flow_pred_ablation}

\begin{table}[h]
\centering\small
\caption{Ablation study of $\Psi$-0.6 models: effect of pMotion, CFG scale, and sequence length on optical flow prediction.
$n=622$.}
\label{tab:flow_epe_ablation}
\resizebox{\textwidth}{!}{%
\begin{tabular}{lccccc|c|c}
\toprule
Method & Acc$\uparrow$ & rEPE$_{\mathrm{move}}$ $\downarrow$ & rEPE$_{\mathrm{move}}$ med $\downarrow$ & Blank $\downarrow$& Dir ($^\circ$) $\downarrow$& $\Delta$PSNR $\uparrow$& Oracle Input \\
\midrule
$\Psi$-0.6 Default seq 32 & 31.8\% & 0.635 & 0.639 & 27.5\% & 14.1 & \textbf{+1.60}	& --\\
+ pMotion          & 44.2\% & 0.577 & 0.474 & 34.6\% & \textbf{6.6} &\underline{+1.59}& --\\
+ pMotion + CFG1.5 & 54.0\% & 0.463 & 0.363 & 5.6\% & \underline{7.7} & +1.49& --\\
+ pMotion + CFG2.0 & \textbf{56.4\%}& \textbf{0.454} & \underline{0.356} & \underline{0.3\%} & 8.1 &+0.94&-- \\
+ pMotion + CFG2.5 &\textbf{56.4\%} & 0.463 & \textbf{0.347} & \textbf{0.0\%} & 9.1 & +0.25 & --\\
+ pMotion + CFG1.5 + seq 128 & 54.2\% &\underline{0.462}& 0.364 & 5.6\% & 7.8 & +1.45& -- \\
+ pMotion + CFG1.5 + seq full &54.2\%&\underline{0.462}& 0.364 & 5.3\% & 7.8 & +1.39 & --\\
\bottomrule
\end{tabular}%
}

\end{table}

To improve text-to-flow predictions, $\Psi$-0.6 further employs two powerful decoding techniques. Critically, these are enabled by $\Psi$-0.6's fine-grained control and probabilistic predictions, attributes that are absent from many other world models. 
First, before generating flow, $\Psi$-0.6 computes the probability of motion (``\emph{pMotion}'') at every patch with a single feedforward pass. Then, $\Psi$-0.6 generates flow predictions sequentially from patches most likely to move to least likely. The intuition behind this is: \yifan{in \dataset{\dname}, motion typically involves a single, often small object, so predicting flow for the patches most likely to move first helps the model focus on the relevant region.} 
Second, our model uses a form of classifier-free guidance (CFG)~\citep{ho2022cfg}: at every generated token, compare the flow prediction logits with and without conditioning on the text instruction, then amplify the difference -- this amplifies the effect of the instruction.

Table~\ref{tab:flow_epe_ablation} shows that pMotion raises accuracy from $31.8\%$ to $44.2\%$ and reduces directional error from $14.1^\circ$ to $6.6^\circ$. CFG at scale 1.5 further improves accuracy to $54.0\%$, lowers rEPE$_{\mathrm{move}}$ to $0.463$, and reduces the blank rate to $5.6\%$. Stronger guidance further improves flow metrics but compromises reconstruction: scale 2.5 eliminates blanks while reducing reconstruction gains from $+1.49$ to $+0.25$ dB. Decoding beyond 32 tokens yields no further improvement, likely due to \dname's sparse, single-object motions. These results favor pMotion with moderate CFG as the best overall configuration, without requiring longer decoding.



\subsection{Question--answer data for stream selection}
\label{sec:pwpn_data}
The stream-selection policy (\S\ref{sec:rl}) is trained and evaluated on videos from four sources: EgoDex~\citep{hoque_egodex_2026}, Something-Something~v2 (SSv2)~\citep{goyal_something_2017}, PE~\citep{bolya_perception_2025}, and OpenVid-1M~\citep{nan_openvid-1m_2025}. None of these datasets provides question--answer pairs. Each provides a caption or label per clip, which we convert into open-ended question--answer pairs. SSv2 labels (e.g., ``covering glass with handkerchief'') and EgoDex task descriptions (e.g., ``Place the lid onto the cup on the wooden table.'') are short and formulaic, so we convert them deterministically. PE and OpenVid captions are longer free-form descriptions. For these, we prompt Qwen2.5-7B-Instruct~\citep{qwen_qwen25_2025} with the caption alone, without the video, to write one to three question--answer pairs. The prompt asks for temporal, spatial, causal, descriptive, action, or narrative questions that are answerable from the caption and add no new information.

\paragraph{Unseen evaluation benchmarks.} We evaluate transfer of the stream-selection policy on four multiple-choice video question-answering benchmarks, none of which appears in training: HanDyVQA \citep{tateno_handyvqa_2026}, which probes fine-grained hand--object interaction dynamics in egocentric video; TGIF-QA~\citep{jang_tgif-qa_2017}, which tests spatio-temporal reasoning such as repeated actions and state transitions; STAR~\citep{wu_star_2024}, which tests situated reasoning about interactions, sequences, predictions, and feasibility in real-world videos; and EgoPlan-Bench2~\citep{qiu_egoplan-bench2_2025}, which asks for the next action toward a stated goal in everyday egocentric tasks. The policy selects streams from the video and question alone, without access to the answer options.

\clearpage
\newpage

\section{Supplementary Text}
\label{sec:supplementary_text}

\subsection{Preliminaries on Probabilistic Structure Integration (PSI)}
\label{sec:preliminaries_psi}

$\Psi$-0.6 builds on Probabilistic Structure Integration (PSI)~\citep{kotar_world_2025}, inheriting two further properties that make its prompting interface unusually flexible: \emph{fine-grained control} over individual patches and \emph{probabilistic prediction}, which allows unspecified content to remain open to multiple plausible continuations.

\textbf{Fine-grained control.}
$\Psi$-0.6 represents visual fields as independently encoded spatial patches, using \emph{local patch quantization} together with explicit \emph{pointer tokens}. Each RGB, flow, or depth patch is represented by content tokens describing \emph{what} the patch contains and a pointer token specifying its modality, spatial location, and temporal location---that is, \emph{where} and \emph{when} the content belongs. Because content and location are represented separately, a prompt can address only selected regions of a visual field, leave the remaining regions unspecified, or relocate existing content by changing its pointer while preserving its content tokens. This enables fine-grained conditioning and targeted scene edits. For example, in Fig.~\ref{fig:architecture}B, moving a patch on the upper part of a can downward while keeping the base fixed specifies a compression, whereas moving it upward without fixing the base specifies a lift. The same representation also allows patches to be generated in arbitrary spatial order, including sequentially (autoregressively), in parallel, or through a combination of both.

\textbf{Probabilistic prediction.}
Prompts specify constraints rather than a single deterministic target. For any partially specified scene, $\Psi$-0.6 represents a distribution over compatible continuations, allowing the same prompt to yield alternative predictions through sampling.

\subsection{Supplementary Discussion}

\paragraph{Unifying reasoning and imagination.}
The value of a unified predictive model becomes particularly clear when reasoning requires considering a world that has not yet been observed. A growing line of work augments language-based reasoners with external visual generators or world models. Mindjourney~\citep{yang2026mindjourney} invokes a frozen Stable Virtual Camera~\citep{zhou2025stable}; SandboxVLM introduces expert models and 3D reconstruction~\citep{liu2026abstract}; Thinking with Novel Views~\citep{zhang_thinking_2026} uses an external ``Painter''; APC~\citep{lee_perspective-aware_2025} chains multiple foundation models; and AVIC~\citep{yu_when_2026} learns when to invoke a world model. These approaches demonstrate the value of visual imagination, but retain a separation between the system that reasons and the system that imagines. World2VLM~\citep{zhang_world2vlm_2026} finds benefits from distilling imagined experience into a VLM rather than relying only on runtime coupling. $\Psi$-0.6 pushes this integration further: the model that interprets an observation is also the model that predicts its geometry, motion, and possible transformations. Imagined states are generated and consumed within the same model and token interface as the rest of the reasoning process. When the reasoner and the imaginer are the same model, imagination becomes a first-class reasoning operation rather than a separately invoked tool.
\clearpage
\section{Model and Training Details for Reproducibility}
\label{sec:technical_details}
\suppressfloats[t]

We describe the implementation of the unified model, its prompting interface, the training of the stream-selection policy, and the supervised training recipe, following the organization of \S\ref{sec:method}. The model inherits the patch-based representation and conditional-prediction approach of PSI~\citep{kotar_world_2025}; here we specify the tokenization, transformer, sequence construction, and optimization needed to implement these mechanisms. The base recipe describes $\Psi$-0.6 after approximately 1.2 million supervised training steps, followed by context-length extension. The initialization of the stream-selection policy and its answer scorer is described in \S\ref{sec:technical_adaptive}.

\subsection{Unified Multimodal Model}
\label{sec:technical_model}

\subsubsection{A common token and position interface}

Each sequence element represents one image, a depth map, an optical-flow field, a relative camera pose, or a text segment. We serialize elements into a stream of discrete token IDs. Each token also carries a position $(x,y,t,c)$: two spatial coordinates, a temporal index, and a channel specifying the representation and, where applicable, its code component. The transformer input is the sum of a token embedding and a channel embedding. Spatial and temporal positions enter attention through rotary embeddings.

For a visual patch, serialization emits a fixed pointer token followed by the patch's content codes. The pointer has token ID 65,000; its position identifies the requested patch, and its channel is the first content channel of that modality. The same pointer ID is reused across locations and modalities. Pointers supply the location as conditioning information, and prediction loss is applied to the supervised content codes.

The transformer has one 65,536-entry token embedding and output vocabulary. Code indices overlap across modalities: for example, a visual code and a text token can have the same integer ID, with their channel embeddings distinguishing their roles. Table~\ref{tab:technical_tokens} summarizes the representations used in the supervised training mixture for $\Psi$-0.6.

\begin{table}[tbp]
\centering
\small
\renewcommand{\arraystretch}{1.12}
\begin{tabular}{@{}p{0.16\linewidth}p{0.25\linewidth}p{0.49\linewidth}@{}}
\toprule
Element & Serialized size & Content representation \\
\midrule
RGB patch & 1 pointer + 4 codes & Learned codes for a $16\!\times\!16$ pixel patch. \\
Depth patch & 1 pointer + 2 codes & Learned codes for normalized inverse depth on the same patch grid. \\
Flow patch & 1 pointer + 2 codes & Learned codes for two-dimensional displacement; channels also encode the temporal gap. \\
Camera pose & 1 pointer + 8 values & Two fields of view, three axis--angle rotation components, and three translation components. \\
Text segment & 1 pointer + text tokens & GPT-2 byte-pair encoding, followed by an end-of-text token. \\
\bottomrule
\end{tabular}
\caption{Token representations. Pointer tokens count toward sequence length but do not receive prediction loss. Patch sizes are in image pixels, whereas the model's spatial coordinates index the patch grid.}
\label{tab:technical_tokens}
\end{table}

\subsubsection{Visual quantization and geometric labels}

\paragraph{Local patch codes.} RGB, depth, and flow use separately trained, frozen patch quantizers. Their encoders apply a Haar wavelet transform within each $16\!\times\!16$ patch, project the patch to 512 features, and process it with three residual blocks using $1\!\times\!1$ convolutions. Each residual block uses channel-wise layer normalization (LayerNorm, $\epsilon=10^{-5}$), a hidden width of 1,024, and sigmoid linear unit (SiLU) activations. Encoding is therefore local to each patch. Finite scalar quantization (FSQ) maps six scalar components, with $(8,8,8,5,5,5)$ levels, to one of $8^3 5^3=64,000$ discrete indices. Quantizer training uses a straight-through estimator (STE) for rounding. There are four such codes per RGB patch and two per depth or flow patch. The first code has an auxiliary coarse-reconstruction objective. Decoders use nine residual blocks with $3\!\times\!3$ convolutions on the code grid followed by an inverse Haar wavelet transform, so reconstruction can combine information across neighboring patches.

The RGB quantizer was trained on ImageNet and OpenImages with an $L_1$ reconstruction term, a DINOv2 feature term, and a coarse $L_1$ term weighted by 0.01. The flow quantizer uses squared reconstruction error and a coarse squared error term weighted by 0.01. These quantizers are trained independently and frozen during autoregressive transformer training, which uses precomputed visual codes.

\paragraph{Depth and camera geometry.} The corpus preprocessing pipeline uses Depth Anything 3 (DA3) to estimate depth, camera intrinsics, and camera poses from video. Depth is converted to inverse depth, normalized to a clip's inverse-depth range, and quantized as a three-channel replicated scalar image. Reconstruction uses the stored per-clip normalization bounds to convert decoded values back to depth.

\paragraph{Optical flow.} The flow preprocessing pipeline estimates displacement with AllTracker in two-frame mode. Horizontal and vertical displacement are expressed in pixels at the preprocessing resolution, divided by 40 before quantization, and multiplied by 40 after decoding. Each stored field has an associated frame gap. Training sequences use only frame pairs for which the required flow field is available. Flow positions carry the source time, while a gap-dependent pair of channels distinguishes the two content codes and the temporal interval.

\paragraph{Camera-pose discretization.} Given camera-to-world matrices $T_i$ and $T_j$, the camera element represents $T_i^{-1}T_j$. Horizontal and vertical fields of view are rounded to integer degrees. The rotation is expressed as a three-component axis--angle vector in degrees; each component is clipped to $[-180,180]$, rounded, and shifted by 180. Translation components are expressed in centimeters, clipped to $[-1000,1000]$, rounded, and shifted by 1000. This deterministic discretization produces eight integers occupying fixed component channels. The sequence notation identifies the source and destination frames; camera tokens use spatial coordinates $(0,0)$ and the source frame's temporal coordinate.

\subsubsection{Text representation}

Text uses GPT-2 byte-pair encoding (BPE) and end-of-text (EOT) ID 50,256. A pointer precedes each text segment. To use the same spatial-position interface as visual tokens, text is laid out in rows of width 100: token position $k$ has $x=k\bmod 100$ and $y=\lfloor k/100\rfloor$. Its temporal index associates the segment with a frame or clip. Questions and answers share the base text channel and are distinguished by their position in the causal sequence. The stream-selection policy uses a distinct reasoning channel and text delimiters around the emitted notation.

\subsubsection{Transformer architecture}

The model is a decoder-only transformer with approximately 8.05 billion parameters. Each block uses a pre-norm layout, applying root mean square normalization (RMSNorm) before grouped-query attention (GQA) with a causal mask and before a Swish-gated linear unit (SwiGLU) feed-forward network, with residual connections around both operations. Attention uses scaled dot products with factor $1/\sqrt{128}$. The output projection shares weights with the input token embedding. Within each 128-dimensional attention head, three-axis rotary position embeddings (RoPE) allocate 48 dimensions to $x$, 48 to $y$, and 32 to $t$, with frequency base 100. Explicit coordinates preserve a patch's identity when its place in the serialized sequence changes, and the causal mask determines which earlier tokens it can attend to.

\begin{table}[tbp]
\centering
\small
\renewcommand{\arraystretch}{1.12}
\begin{tabular}{@{}ll@{}}
\toprule
Setting & Value \\
\midrule
Transformer blocks / hidden width & 32 / 4,096 \\
Query heads / key--value heads & 32 / 8 \\
Head dimension / SwiGLU hidden width & 128 / 16,384 \\
Token vocabulary / channel embeddings & 65,536 / 384 \\
Base training sequence length & 4,096 tokens \\
Normalization & RMSNorm, $\epsilon=10^{-5}$ \\
Input--output token embeddings & Tied \\
Dropout / stochastic depth / linear biases & 0 / 0 / disabled \\
\bottomrule
\end{tabular}
\caption{Architecture of $\Psi$-0.6. The channel table reserves capacity beyond the modalities used in this recipe.}
\label{tab:technical_architecture}
\end{table}

\subsection{Generic Prompting Interface: Streams-of-Thought}
\label{sec:technical_prompting}

\subsubsection{Notation and sequence construction}

A notation lists the elements of a sequence in causal order. Node elements refer to one time index, as in $\mathrm{RGB}_0$ or $D_0$; edge elements refer to a pair, as in $C_{0\to1}$ or $F_{0\to1}$. Here $D$, $C$, and $F$ denote depth, relative camera pose, and optical flow, respectively. At inference, an arrow separates supplied context from requested outputs. For example, $\mathrm{RGB}_0,C_{0\to1}\rightarrow\mathrm{RGB}_1$ requests a new image conditioned on a source image and camera motion, whereas $\mathrm{RGB}_0,\mathrm{RGB}_1\rightarrow C_{0\to1}$ requests the relative camera pose. Longer streams concatenate these operations, making generated content available to later predictions.

Training uses complete sequences sampled from a finite, weighted set of notations for each dataset group. The temporal sampler maps the notation's indices to available frames and geometric edges in a clip. Each element is then loaded, allocated a token budget, and serialized. Examples from the instruction mixture include $\mathrm{RGB}_0,C_{0\to1},F_{0\to1},\mathrm{RGB}_1$ for geometric video modeling; $\mathrm{RGB}_0,D_0,T_0$ for image description; and $Q,\mathrm{RGB}_0,D_0,A$ for answering an image question with depth, where $T_0$ is a caption and $Q$ and $A$ are question and answer text. Video-question sequences can place the question before or after the images, and geometric representations can be interleaved with images or placed between the question and answer.

\paragraph{Token allocation and patch order.} During base training, the 4,096-token budget includes pointers and conditioning tokens. Text and fixed-length camera elements are accounted for first. The remaining budget is distributed approximately uniformly in patch count across image, depth, and flow elements, subject to their available patch counts. The allocator accounts for five tokens per RGB patch and three per depth or flow patch, with stochastic remainder allocation and a final length adjustment to meet the sequence budget. Within each visual element, patches are randomly permuted and the allocated subset is retained; text retains its linguistic order. Short sequences are padded with masked positions.

\paragraph{Element selection and supervision.} Sampling different notations exposes the model to different conditioning sets and prediction orders. Each dataset group's loss mask specifies the conditioning elements and supervised targets. The first RGB element usually serves as context; the final global-caption group also supervises it, supporting text-first image prediction. Question-answer groups supervise answer text, with their image, depth, flow, pose, and question tokens providing context. The group masks and mixture weights are described in \S\ref{sec:technical_recipe}.

\subsubsection{Generating and composing elements}

Generation begins by serializing the supplied elements and constructing a template for the requested element. The template supplies pointer tokens, positions, and channels, while leaving content slots to be sampled from the transformer's output distribution. Content IDs are sampled within the configured range of the target modality. The resulting codes can be retained as context for the next element or decoded into a visual representation.

To accelerate visual inference, the decoder combines an autoregressive patch prefix with parallel decoding of the remaining patches. It first generates a chosen number of patches sequentially, caching attention keys and values to establish shared structure. The remaining patches are processed concurrently, with content codes within each patch generated sequentially. Each parallel patch attends to the supplied context, the completed prefix, and its own preceding content codes. This reduces the number of sequential decoding steps needed to complete a visual output. The sequential budget is measured in patches.

Temperature, top-$k$, nucleus threshold, patch ordering, and sequential patch budget control inference. Generic generation defaults to temperature 1, top-$k$ 1,000, nucleus threshold 1, and 32 sequential patches. Text is decoded in token order and truncated at its first end-of-text token within a specified output budget.

\paragraph{Guidance and targeted patch ordering.} Classifier-free guidance (CFG) combines conditional and reference logits before sampling:
\begin{equation}
 z_{\mathrm{guided}}=z_{\mathrm{ref}}+
 s\bigl(z_{\mathrm{cond}}-z_{\mathrm{ref}}\bigr).
\end{equation}
The reference context is formed by omitting selected elements, such as a motion instruction, while retaining other context such as the input image. For sparse-motion outputs, the probability that a patch moves can also be estimated from the first flow-code distribution. With $\mathcal Z$ denoting the seven code indices identified as no-motion codes, the ordering score is $1-\sum_{z\in\mathcal Z}p(z)$. Decoding high-scoring patches first allocates the sequential budget to likely moving regions. Guidance and motion-based patch ordering provide task-specific control over inference.

\subsection{Training Adaptive Streams-of-Thought}
\label{sec:technical_adaptive}

The adaptive-stream extension trains a policy to emit a notation from a video and question. The policy learns to select and order depth, camera-motion, and flow representations. These selected elements are assembled with the video and question as context for an answer scorer, whose answer loss supplies the policy's training signal.

\paragraph{Policy initialization by scored search.} For each training video-question pair, 64 candidate notations are scored using the model's teacher-forced mean cross-entropy on the reference answer. Candidate selection considers the ten lowest-loss notations and prioritizes improvements over RGB-only answering. Sampling weights are proportional to $\exp(-\ell/0.5-0.01m)$, where $\ell$ is answer loss and $m$ is the number of selected elements. The model is then supervised on the notation and answer tokens, with the video and question as context. These examples are mixed with replay of the vision-language training data. The four notation-training sources are EgoDex, Something-Something v2, PE-Video, and OpenVid.

\paragraph{On-policy optimization.} Subsequent optimization samples $K=16$ notations per video-question pair and updates low-rank adaptation (LoRA) adapters on the initialized policy. Let $\ell(a;x,q,y)$ be the scorer's mean answer-token loss when the notation $a$ selects additional context for video $x$, question $q$, and reference answer $y$. Reward favors lower $\ell$; subtracting an RGB-only baseline gives the equivalent within-input objective
\begin{equation}
 r(a;x,q,y)=\ell(\varnothing;x,q,y)-\ell(a;x,q,y).
\end{equation}
Because the baseline is shared by a group, it cancels when forming $A_i=r_i-K^{-1}\sum_j r_j$. We use Dr.~GRPO, a variant of group relative policy optimization (GRPO), with group-mean reward centering, scaling by a global running root-mean-square statistic, and advantage clipping to $[-3,3]$. It averages reward measurements over three paired random seeds, uses a fixed maximum-output-length normalizer for notation-token log-probabilities, and regularizes the policy against its initialization. The policy receives the video and question, while the reference answer supplies the training reward.

\paragraph{Policy and scorer initialization.} The policy and answer scorer are initialized from instruction-trained models. Policy distillation runs for approximately 40k steps. Subsequent LoRA training uses learning rate $3\!\times\!10^{-5}$, Kullback--Leibler (KL) regularization coefficient 0.02, and ranks 128, 256, or 512, with longer runs trained for approximately 6k steps.

\subsection{Training Recipe: Pretraining, Mid-training, and Post-training}
\label{sec:technical_recipe}

\subsubsection{Supervised objective and sampling}

Each dataset group specifies data sources, notation weights, a token allocator, and a supervision mask. Group weights are normalized to sampling probabilities, determining each group's share of sampled sequences. Within a group, eligible files are drawn from the pooled sources, and a notation is sampled according to its normalized weight.

All stages optimize next-token cross-entropy. Pointer, padding, and group-specific conditioning positions are excluded from the loss. Let $\mathcal T_b$ be the supervised positions in sample $b$, and let $\ell_{bi}$ be the next-token negative log-likelihood at position $i$. Instruction training uses
\begin{equation}
 \mathcal L_{\mathrm{instruction}}=
 \frac{1}{B}\sum_{b=1}^{B}
 \frac{\sum_{i\in\mathcal T_b}\ell_{bi}}
 {\max(1,|\mathcal T_b|)}.
 \label{eq:technical_loss}
\end{equation}
This gives each sequence equal weight despite differences in the number of supervised tokens. The earlier pretraining and alignment stages instead average over supervised tokens across the batch, $\sum_b\sum_{i\in\mathcal T_b}\ell_{bi}/\sum_b|\mathcal T_b|$. Camera-pose tokens initially serve as conditioning information and are excluded from the loss; they become supervised prediction targets during caption alignment.

\subsubsection{Training stages and data mixtures}

Table~\ref{tab:technical_stages} summarizes the supervised training stages for $\Psi$-0.6. The initial video stage trains RGB-only prediction. The next stage adds depth and flow prediction and camera conditioning. Language is introduced with text-only data, followed by image/video caption alignment, then instruction training with question answering and text-conditioned visual prediction. Earlier predictive tasks remain in the mixture.

\begin{table}[tbp]
\centering
\small
\renewcommand{\arraystretch}{1.12}
\begin{tabular}{@{}p{0.23\linewidth}p{0.22\linewidth}p{0.44\linewidth}@{}}
\toprule
Stage & Steps ($\approx$) & Sequence mixture \\
\midrule
RGB pretraining & 0--500k & Video prediction with 2--8 RGB frames. \\
Geometric integration & 500--700k & RGB, depth, flow, and camera-conditioned video sequences. \\
Language introduction & 700--900k & 50\% vision; 50\% FineWeb-Edu. \\
Caption alignment & 900k--1.03M & 25\% each: vision, FineWeb-Edu, global captions, and local captions. \\
Instruction training & 1.03--1.10M & Captioning, question answering, text control, and predictive replay; DCLM replaces FineWeb-Edu. \\
Instruction refinement & 1.10--1.20M & More video-question training and revised caption/geometry mixtures (Table~\ref{tab:technical_mixture}). \\
\bottomrule
\end{tabular}
\caption{Supervised training stages for $\Psi$-0.6, with approximate cumulative step boundaries. Every stage uses batches of 512 sequences and a 4,096-token sequence budget. The first approximately 180k steps use TPU v6e; subsequent stages use TPU v5e.}
\label{tab:technical_stages}
\end{table}

FineWeb-Edu and DCLM are used at different training stages. FineWeb-Edu is the text-only source during language introduction and caption alignment. DCLM-baseline is the text-only source in both instruction stages. The model's depth, hidden width, and token vocabulary remain unchanged.

\paragraph{Final instruction mixture.} The final stage has nine groups, with approximate sampling proportions shown in Table~\ref{tab:technical_mixture}. Relative to the preceding instruction stage, video-question sampling increases from approximately 5\% to 15\%, and local-caption sampling decreases from approximately 15\% to 10\%. Global-caption notations expand from 80 to 151 and video-question notations from 26 to 42.

\begin{table}[tbp]
\centering
\small
\renewcommand{\arraystretch}{1.12}
\begin{tabular}{@{}p{0.23\linewidth}rp{0.52\linewidth}@{}}
\toprule
Group & Share ($\approx$) & Data and supervision \\
\midrule
Visual prediction & 20\% & Video corpus; content loss after the first RGB element. \\
Text-only & 20\% & DCLM-baseline; language-modeling loss. \\
Image captions & 5\% & COCO train and PixMo-Cap; captions and, in most notations, depth targets. \\
Image questions & 5\% & PixMo-Cap-QA, VQA v2, LLaVA-Instruct-150K, and the naturalistic Cambrian subset; answer-only loss. \\
Global video captions & 20\% & Clip-level captions with visual/geometry sequences; all content elements supervised. \\
Video questions & 15\% & OpenVid, PE-Video, and SSv2 question--answer data; answer-only loss. \\
Global camera control & 1\% & CO3D and MVImgNet; text-conditioned camera/visual prediction and captioning. \\
Local video captions & 10\% & Time-indexed captions interleaved with video and geometry; first RGB masked. \\
Local camera control & 1\% & SpatialVid; time-indexed text and camera/visual sequences. \\
\bottomrule
\end{tabular}
\caption{Approximate sampling proportions during instruction refinement, rounded to whole percentages. All groups mask pointers and padding.}
\label{tab:technical_mixture}
\end{table}

\subsubsection{Data sources and text construction}

\paragraph{Video corpus.} The visual-prediction mixture combines 24 video sources spanning egocentric recordings (Aria, EgoDex, Ego-Exo4D, and Egocentric10K), robot interaction (Bridge, DROID, FMB, and Language-Table), multiview scenes and objects (CO3D, DL3DV, MVImgNet, RealEstate10K, ScanNet, ToyBox, and ZeroShape), human activity (AIST, CelebV-HQ, Kinetics-700, Motion-X, and Something-Something v2), and web video (PE-Video, SpatialVid, YouTube, and a hand-selected collection). The initial stage uses RGB from these sources. Geometric integration joins the available RGB, depth, flow, and pose codes; sources without a modality support only the compatible notations.

\paragraph{Captions and instruction data.} Caption alignment adds global text from COCO, EgoDex, LVD-2M, OpenVid, PE-Video, and Bridge, together with time-indexed text from Ego-Exo4D and AgiBot. Instruction training retains global captions from EgoDex, OpenVid, PE-Video, Bridge, SSv2, and DROID, and adds Qwen-captioned material from the hand-selected, SpatialVid, and YouTube collections. The local-caption group uses Ego-Exo4D, AgiBot, EPIC-KITCHENS clips, and local captions from those three additional video collections. Global captions can precede or follow visual sequences, while local captions are interleaved with their associated times. This trains both description and text-conditioned prediction using the same token interface.

Image instruction data are listed in Table~\ref{tab:technical_mixture}.

\paragraph{Question--answer construction.} The global video-question group uses question--answer pairs constructed from OpenVid, PE-Video, and SSv2 text. The SSv2 conversion uses fixed action-question templates and rewrites the source action label into an answer. The caption rephrasing pipeline for PE-Video and OpenVid prompts a language model with the caption text alone to produce one to three question--answer pairs, then concatenates multiple pairs into one question field and one answer field. Its prompt requires that answers be supported by the caption and forbids adding information.

\subsubsection{Optimization and infrastructure}

Supervised training uses AdamW, Adam with decoupled weight decay, with $\beta_1=0.9$, $\beta_2=0.95$, $\epsilon=10^{-8}$, weight decay 0.1 on matrix parameters, and global gradient norm clipping at 1.0. Normalization gains are not weight-decayed. The learning rate warms up linearly for the first 2,000 steps to $3\!\times\!10^{-4}$, then remains constant until the final 50,000 supervised training steps, when it decreases linearly toward zero.

The base training stages in Table~\ref{tab:technical_stages} use a global batch of 512 sequences, with 4,096 positions per sequence. Training uses bfloat16 (BF16) autocasting, full-precision (FP32) master weights and optimizer state, and activation checkpointing with recomputation at transformer-block boundaries. The implementation uses PyTorch/XLA single-program, multiple-data (SPMD) execution with fully sharded data parallelism (FSDP) and a causal FlashAttention kernel.

Over approximately 1.2 million supervised training steps, the sequence budget corresponds to
\begin{equation}
 1.2\!\times\!10^6\times512\times4096
 \simeq 2.5\!\times\!10^{12}
\end{equation}
processed sequence positions, counting all positions in the training sequences. The standard $6ND$ estimate with $N\simeq8.05\!\times\!10^9$ parameters gives approximately $1.2\!\times\!10^{23}$ FLOPs for the base transformer's supervised training.

\subsubsection{Context-length extension}
\label{sec:technical_context_extension}

For tasks requiring longer multimodal sequences, we continue training instruction-trained models with sequence budgets of 16,384 or 32,768 tokens. We retain the transformer architecture, modality tokenizers, spatial--temporal position representation, and next-token prediction objective. The larger budget accommodates more visual patches and longer streams of interleaved images, geometry, and text. During this phase, context parallelism distributes positions within each sequence across devices to manage memory use.

\clearpage
\subsection{Technical Glossary and Hyperparameter Reference}
\label{sec:technical_glossary}

Tables~\ref{tab:technical_glossary_model}--\ref{tab:technical_glossary_policy} collect the named methods and numerical settings used throughout this appendix. They cover the model and its representations, supervised training, and the decoding and adaptive-policy procedures.

\begin{table}[htbp]
\centering
\small
\renewcommand{\arraystretch}{1.12}
\begin{tabular}{@{}p{0.29\linewidth}p{0.67\linewidth}@{}}
\toprule
Component / term & Method and setting \\
\midrule
Transformer & Decoder-only; approximately 8.05B parameters; 32 blocks; hidden width 4,096. \\
RMSNorm & Root mean square normalization; pre-norm residual blocks; $\epsilon=10^{-5}$. \\
GQA & Grouped-query causal self-attention; 32 query heads, 8 key--value heads; head dimension 128. \\
Attention scaling & Scaled dot-product attention; logit scale $1/\sqrt{128}$. \\
FlashAttention & Memory-efficient causal attention kernel used in TPU training. \\
SwiGLU & Swish-gated linear unit feed-forward network; hidden width 16,384. \\
Three-axis RoPE & Rotary position embeddings on $(x,y,t)$; dimensions $(48,48,32)$ per head; frequency base 100. \\
Weight tying / channels & Shared input embedding and output projection; vocabulary 65,536; 384 channel embeddings. \\
Transformer regularization & Dropout 0; stochastic depth 0; linear biases disabled. \\
\midrule
Haar wavelet transform & Patch-local transform on $16\!\times\!16$ pixels; inverse transform after decoding. \\
Patch autoencoder & Encoder: 3 residual blocks with $1\!\times\!1$ convolutions. Decoder: 9 blocks with $3\!\times\!3$ convolutions. Feature width 512; residual hidden width 1,024. \\
LayerNorm / SiLU & Quantizer residual blocks use channel-wise layer normalization ($\epsilon=10^{-5}$) and sigmoid linear unit activations. \\
FSQ & Finite scalar quantization; levels $(8,8,8,5,5,5)$; 64,000 indices per code; 4 codes per RGB patch, 2 per depth or flow patch. \\
STE & Straight-through estimator for rounding during quantizer training. \\
Pointer tokens & One supplied pointer per patch, camera element, or text segment; token ID 65,000. \\
GPT-2 BPE / EOT & Byte-pair encoding for text; end-of-text ID 50,256; spatial row width 100 tokens. \\
DA3 & Depth Anything 3 supplies depth, camera intrinsics, and poses; depth codes represent normalized inverse depth. \\
AllTracker & Two-frame optical flow; displacement divided by 40 before quantization and multiplied by 40 after decoding. \\
Camera discretization & Relative pose $T_i^{-1}T_j$; integer-degree fields of view and axis--angle rotation; integer-centimeter translation. Rotation components clipped to $[-180,180]$; translation to $[-1000,1000]$. \\
\bottomrule
\end{tabular}
\caption{Technical glossary: architecture, tokenization, and geometric representations. Transformer RMSNorm and quantizer LayerNorm apply to their respective networks.}
\label{tab:technical_glossary_model}
\end{table}

\begin{table}[htbp]
\centering
\small
\renewcommand{\arraystretch}{1.12}
\begin{tabular}{@{}p{0.29\linewidth}p{0.67\linewidth}@{}}
\toprule
Component / term & Method and setting \\
\midrule
Next-token CE & Cross-entropy on supervised content tokens; token-mean reduction in pretraining/alignment and sequence-mean reduction in instruction training. \\
AdamW & Adam with decoupled weight decay; $\beta_1=0.9$, $\beta_2=0.95$, $\epsilon=10^{-8}$. \\
Weight decay & 0.1 on matrix parameters; 0 on normalization gains. \\
Gradient clipping & Global gradient norm capped at 1.0. \\
Learning-rate schedule & Peak $3\!\times\!10^{-4}$; 2,000-step linear warmup; constant phase; linear decay toward zero over the final 50,000 steps. \\
Batch / training duration & 512 sequences per global batch; approximately 1.2M supervised steps for the base model. \\
Context length & 4,096 positions during base training; 16,384 or 32,768 during context extension. \\
Token allocation & Approximately uniform visual patch counts per element within the sequence budget; random visual patch order. \\
BF16 / FP32 & Bfloat16 autocasting; 32-bit floating-point master weights and optimizer state. \\
Activation checkpointing & Recompute transformer-block activations during backpropagation. \\
SPMD / FSDP & Single-program, multiple-data execution and fully sharded data parallelism in PyTorch/XLA. \\
Context parallelism & Distribute sequence positions across devices during context-length extension. \\
TPU hardware & TPU v6e for the first approximately 180k steps; TPU v5e for subsequent base-training stages. \\
\midrule
RGB quantizer objective & Mean absolute error (MAE/$L_1$), DINOv2 feature loss with weight 1, and coarse $L_1$ loss with weight 0.01. \\
Flow quantizer objective & Mean squared error (MSE) and coarse MSE with weight 0.01. \\
\bottomrule
\end{tabular}
\caption{Technical glossary: supervised optimization and training infrastructure. Optimizer and schedule settings refer to the transformer; the final two rows describe the independently trained visual quantizers.}
\label{tab:technical_glossary_training}
\end{table}

\begin{table}[htbp]
\centering
\small
\renewcommand{\arraystretch}{1.12}
\begin{tabular}{@{}p{0.29\linewidth}p{0.67\linewidth}@{}}
\toprule
Component / term & Method and setting \\
\midrule
KV caching & Key--value caching for autoregressive patch decoding. \\
Sequential / parallel decoding & Inference speedup using an autoregressive patch prefix followed by concurrent decoding of remaining patches; codes within each patch are generated sequentially. \\
Sampling controls & Temperature, top-$k$, and nucleus (top-$p$) sampling; generic defaults $T=1$, $k=1{,}000$, $p=1$, with 32 sequential patches. \\
CFG & Classifier-free guidance; combine reference and conditional logits as $z_{\mathrm{ref}}+s(z_{\mathrm{cond}}-z_{\mathrm{ref}})$. \\
Motion-based patch order & Prioritize flow patches by $1-\sum_{z\in\mathcal Z}p(z)$, where $\mathcal Z$ contains seven no-motion codes. \\
\midrule
Teacher-forced scoring & Mean reference-answer token cross-entropy used to score candidate notations during policy training. \\
Scored-search distillation & 64 candidates; top-10 loss filtering; sampling weight $\exp(-\ell/0.5-0.01m)$, where $m$ is the selected-element count; approximately 40k training steps. \\
Dr.~GRPO & Variant of group relative policy optimization; $K=16$ notations per input; group-mean reward centering; global running RMS advantage scaling. \\
Advantage / reward settings & Advantage clipping to $[-3,3]$; average reward measurements over 3 paired random seeds; fixed maximum-output-length normalization of notation log-probabilities. \\
LoRA & Low-rank adaptation for policy updates; ranks 128, 256, or 512; learning rate $3\!\times\!10^{-5}$; approximately 6k steps in the longer runs. \\
KL regularization & Kullback--Leibler regularization toward the initialized policy; coefficient 0.02 in the longer LoRA runs. \\
\bottomrule
\end{tabular}
\caption{Technical glossary: decoding controls and adaptive-policy training. Sampling values are the generic predictor defaults; policy settings summarize the procedures in \S\ref{sec:technical_adaptive}.}
\label{tab:technical_glossary_policy}
\end{table}
\clearpage

\end{document}